\documentclass[letterpaper]{article} 
\usepackage{aaai2026}  
\usepackage{times}  
\usepackage{helvet}  
\usepackage{courier}  
\usepackage[hyphens]{url}  
\usepackage{graphicx} 
\usepackage{natbib}  
\usepackage{caption} 
\usepackage{algorithm}
\usepackage{algorithmic}

\usepackage{newfloat}
\usepackage{listings}
\DeclareCaptionStyle{ruled}{labelfont=normalfont,labelsep=colon,strut=off} 
\floatstyle{ruled}
\newfloat{listing}{tb}{lst}{}
\floatname{listing}{Listing}
\usepackage{array}
\usepackage{amsmath}
\usepackage{amssymb}
\usepackage{mathtools}
\usepackage{amsthm}
\usepackage{multirow}
\usepackage{cleveref}
\usepackage{booktabs}

\newtheorem{theorem}{Theorem}

\title{From Perturbation Correction to Geometry-Aware Sampling: Sharpness-Guided Equilibrium Sampling for Balanced Flat Minima in Long-Tailed Learning}

\author {
    Jiaxin Deng\textsuperscript{\rm 1},
    Junbiao Pang\textsuperscript{\rm 1}\thanks{Corresponding authors.}
}
\affiliations {
    \textsuperscript{\rm 1}School of Information Science and Technology, Beijing University of Technology, Beijing, China\\
    dengjiaxin@emails.bjut.edu.cn, junbiao\_pang@bjut.edu.cn
}

\usepackage{bibentry}

\begin{document}

\maketitle

\begin{abstract}
Long-tailed learning couples two sources of poor generalization: head classes dominate training exposure, while under-represented classes often converge to sharper regions of the loss landscape. 
Conventional re-sampling addresses the former without considering geometry, whereas existing long-tailed sharpness-aware minimization (SAM) methods modify losses or perturbations only after biased mini-batches have been drawn. We introduce Sharpness-Guided Equilibrium Sampling (SGS), which treats the sampling distribution as an active control variable for optimization geometry. 
SGS dynamically adjusts subsequent mini-batches by increasing the sampling probability of less frequently sampled classes while suppressing classes with large SAM-induced loss changes, using only cumulative class counts and EMA sharpness estimates obtained from the standard SAM update, without class-wise perturbations or additional backward passes. 
We characterize this sampling process through a continuous-time stochastic differential equation and a sampling-dependent PAC-Bayes analysis, explaining how frequency-sharpness feedback can move training toward a more balanced flatness profile. 
On CIFAR-100 LT with an imbalance ratio of 100, SGS-SAM improves Focal-SAM by 10.85 points in tail accuracy and 3.56 points overall. 
On ImageNet-LT, it improves ImbSAM by 6.59 points on tail classes and 1.20 points overall. 
Its training time is only $1.02\times$ that of vanilla SAM. Beyond these gains, SGS establishes a sampling-side route to loss-landscape control, suggesting that future long-tailed methods can jointly regulate data exposure and optimization geometry rather than treating either as fixed.
\end{abstract}


\section{Introduction}
Deep neural networks have achieved remarkable success in visual recognition tasks such as image classification~\cite{wang2023internimage} and object detection~\cite{wang2024yolov10}. However, their performance often relies on balanced and well-curated training data, whereas real-world visual data commonly follows a long-tailed distribution, with a few head classes containing abundant samples and many tail classes being severely underrepresented. Training with standard empirical risk minimization on such imbalanced data can bias the model toward head classes and degrade overall generalization~\cite{cao2019learning}. To address this issue, existing methods mainly focus on re-sampling~\cite{buda2018systematic}, re-balancing~\cite{wang2023unified,li2025focal}, robust representation learning~\cite{cui2023generalized}, and foundation-model fine-tuning~\cite{shi2024long}. 

Existing approaches address different parts of this problem. Re-sampling and class-prior correction improve tail exposure, but they are geometry-agnostic: oversampling can repeatedly reuse scarce tail examples and amplify gradient variance, whereas undersampling discards informative head data~\cite{buda2018systematic}. Sharpness-Aware Minimization (SAM)~\cite{foretsharpness} and its long-tailed variants, including ImbSAM~\cite{zhou2023imbsam}, CC-SAM~\cite{zhou2023class}, and Focal-SAM~\cite{li2025focal}, instead regulate optimization geometry through class-dependent perturbations or penalties. However, these corrections are applied after a biased mini-batch has been drawn. The original long-tailed distribution therefore continues to decide which classes and geometric regions dominate training. This separation leaves a central question unanswered: can data exposure and optimization geometry be controlled jointly through the sampler itself?

We answer this question with Sharpness-Guided Equilibrium Sampling (SGS), which shifts geometry control from post-sampling perturbation correction to mini-batch construction. SGS reuses the perturbed loss gap already produced by standard SAM as an efficient sharpness signal. After each update, SGS records the cumulative exposure and EMA sharpness of each class, uses these statistics to recompute the sampling probabilities, and constructs the next mini-batches accordingly. Under-sampled classes receive more optimization opportunities, whereas classes with large perturbation responses are temporarily down-weighted to avoid repeatedly emphasizing unstable regions. The resulting mini-batches update the model and produce new frequency and sharpness statistics, which are then used for the next sampling decision. Thus, the sampling distribution evolves jointly with the model rather than remaining fixed throughout optimization, without requiring class-wise perturbations or additional backward passes.

We characterize this adaptive sampling process through continuous-time stochastic dynamics and a sampling-dependent PAC-Bayes analysis. The stochastic view separates the deterministic competition induced by class frequency and sharpness from the fluctuations caused by mini-batch sampling. The PAC-Bayes analysis further identifies conditions under which the resulting distribution improves class balance and yields a tighter generalization bound than ordinary long-tailed SAM. Experiments on CIFAR-LT and ImageNet-LT show consistent improvements for medium and tail classes under severe imbalance, while experiments with foundation-model fine-tuning demonstrate the applicability of SGS to a different adaptation pipeline. Importantly, SGS preserves the standard SAM optimization procedure and introduces only lightweight class-statistic and sampling-probability updates, resulting in nearly the same training cost as vanilla SAM. Our contributions are

\begin{itemize}
\item We propose SGS, a geometry-aware sampling method that uses the cumulative exposure and SAM-induced loss response of each class to update future mini-batch composition. SGS reallocates optimization opportunities before parameter updates without class-wise perturbations or additional backward passes.

\item We connect the discrete sampling process to continuous-time stochastic dynamics and a sampling-dependent PAC-Bayes bound, explaining how frequency-sharpness competition, mini-batch noise, and adaptive class exposure jointly affect long-tailed generalization.

\item We validate SGS on CIFAR-LT, ImageNet-LT, and foundation-model fine-tuning. SGS substantially improves medium- and tail-class performance under severe imbalance while retaining nearly the same training efficiency as vanilla SAM, with only approximately $2\%$ additional training time on CIFAR-LT.

\end{itemize}

\section{Related Works}
\subsection{Long-Tailed Learning}
Long-tailed learning has been widely studied to mitigate the bias induced by imbalanced class distributions. Existing methods can be roughly divided into data re-balancing, data augmentation, loss and logit adjustment, representation learning, decoupled training, and ensemble learning. Data re-balancing methods adjust the sampling frequency of different classes, while augmentation methods synthesize or enrich tail samples to alleviate data scarcity \cite{kang2019decoupling,ren2020balanced,wang2020devil}. Loss and logit adjustment methods incorporate class priors into the training objective, such as margin-based losses \cite{cao2019learning}, balanced softmax \cite{ren2020balanced}, and logit correction \cite{menon2020long}. Representation learning methods improve feature separability for tail classes through contrastive objectives \cite{cui2021parametric}, prototypes \cite{wei2022prototype}, or feature augmentation \cite{hong2022safa}. Decoupled training methods separate representation learning from classifier calibration, showing that balanced classifier learning is crucial for long-tailed recognition \cite{zhong2021improving}.

Although these methods have achieved promising results, most of them are driven by class frequency or class priors. Such quantity-based correction is effective for reducing head-class dominance, but it does not explicitly describe whether the sampled instances lie in sharp regions of the loss landscape. SGES complements existing long-tailed learning paradigms by connecting sampling control with local curvature, dynamically allocating optimization effort according to both class imbalance and sharpness information.

\subsection{Sharpness-Aware Minimization in Long-Tailed Learning} 
SAM \cite{foretsharpness} improves generalization by seeking flat minima, which is particularly relevant in long-tailed learning because rare and difficult classes often suffer from compressed decision regions and poor robustness. Several long-tailed SAM variants have therefore been proposed. ImbSAM \cite{zhou2023imbsam} adopts a coarse-grained strategy by restricting SAM perturbations to tail classes. While this partially protects the tail, it may sharpen the head-class landscape and degrade overall performance when paired with standard rebalancing algorithms. CC-SAM \cite{zhou2023class} introduces class-conditional perturbation radii for fine-grained control, but it requires costly class-wise perturbation computations. Focal-SAM \cite{li2025focal} improves the efficiency-control trade-off by applying focal-like weights to class-wise sharpness penalties, thereby strengthening tail-class smoothness without the burden of extra class-wise backpropagation.

These methods provide important evidence that optimization geometry matters for long-tailed recognition. However, they mainly intervene after mini-batches have been drawn, by changing perturbation directions, perturbation radii, or sharpness penalties. In contrast, SGES shifts the control point from the loss/perturbation rule to the sampling distribution itself. SGES reshapes the mini-batch stream according to both frequency and curvature, without requiring class-wise perturbations or additional backward passes.

\section{Method}
\subsection{Frequency and Sharpness Guided Sampling}
Let $N_c$ be the number of training samples in class $c$, $N=\sum_{j=1}^{C}N_j$, and $P_c^{base}=N_c/N$ be the empirical class prior. In long-tailed training, $P_c^{base}$ makes head classes dominate both gradient updates and SAM perturbations. SGS constructs a dynamic sampling distribution by combining frequency feedback and sharpness feedback.

\textbf{Frequency sampling indicator.} Let $m_c(t)$ denote the cumulative number of sampled instances from class $c$ up to step $t$. We define the empirical sampling share as $p_c(t)=m_c(t)/(\sum_j m_j(t)+\xi)$, where $\xi>0$ is a small constant introduced for numerical stability and to avoid division by zero. A smaller $p_c(t)$ indicates that class $c$ has contributed less often to optimization updates. We therefore define the inverse-frequency indicator as follows:
\begin{equation}\label{equ:inverse_frequency_indicator}
\begin{aligned}
w_c^{freq}=1/(p_c(t)+\xi),
\end{aligned}
\end{equation}
which assigns larger scores to classes with small cumulative sampling shares.

\textbf{Sharpness proxy.} Direct Hessian estimation or class-wise SAM perturbation is expensive. SGS instead reuses the perturbation from the standard SAM step. For instance $x_i$ and model weights $w$, we define:
\begin{equation}\label{equ:sharpness_proxy}
\begin{aligned}
\Delta L_{x_i}=|L_{x_i}(w+\epsilon)-L_{x_i}(w)| .
\end{aligned}
\end{equation}
A large $\Delta L_{x_i}$ indicates high sensitivity to the SAM adversarial direction. A local Taylor expansion gives $L_{x_i}(w+\epsilon)-L_{x_i}(w)\approx \nabla L_{x_i}(w)^{\top}\epsilon+\frac{1}{2}\epsilon^{\top}\nabla^2 L_{x_i}(w)\epsilon$, where the quadratic term captures directional curvature around $w$. Thus, $\Delta L_{x_i}$ serves as an efficient update-aware sharpness proxy.

\textbf{Sharpness sampling indicator.} Although tail classes require additional exposure, repeatedly sampling samples from sharp regions may amplify gradient noise and harm generalization. Therefore, SGS uses the sharpness proxy in Eq.~\eqref{equ:sharpness_proxy} to construct a stable class-level sharpness indicator.

For a mini-batch $B$ at step $t$, let $B_c$ denote the samples of class $c$. When $|B_c|>0$, we compute the batch-level class sharpness as
\begin{equation}
\Delta L_c=\frac{1}{|B_c|}\sum_{x_i\in B_c}\Delta L_{x_i}.
\end{equation}
To reduce the stochasticity of mini-batch estimates, we maintain an exponential moving average for each class:
\begin{equation}
h_c^{(t)}=
\begin{cases}
\beta h_c^{(t-1)}+(1-\beta)\Delta L_c, & |B_c|>0,   \\
h_c^{(t-1)}, & |B_c|=0,
\end{cases}
\end{equation}
where $\beta=0.9$ or $0.99$. The EMA reduces the noise caused by infrequent tail-class observations.

We then normalize the class-wise estimates as $\tilde{h}_c = {h_c}/({\sum_{j=1}^C h_j + \xi})$ and define the sharpness sampling weight by
\begin{equation}
w_c^{\mathrm{sharp}}={1}/({\tilde{h}_c^{\kappa}+\xi}),
\end{equation}
where $\kappa>0$ controls the strength of sharpness feedback and $\xi>0$ avoids division by zero. The inverse form assigns lower sampling weight to classes with larger perturbation responses, thereby preventing the sampler from over-emphasizing sharp and unstable regions.

\textbf{Joint sampling weight.}
The joint class score $\tilde{w}_c$ combines both sampling indicators:
\begin{equation}\label{equ:joint_w_unnorm}
\begin{aligned}
\tilde{w}_c = w^{freq}_c \cdot w^{sharp}_c
\end{aligned}
\end{equation}
We apply mean normalization to ensure numerical stability, yielding the normalized class sampling weight $w_c$:
\begin{equation}\label{equ:class_w_final}
\begin{aligned}
\quad w_c = \frac{\tilde{w}_c}{\frac{1}{C}\sum_{j=1}^C \tilde{w}_j + \xi}
\end{aligned}
\end{equation}
where $C$ represents the total number of classes.

\textbf{Instance-Level Probability Allocation.} The calibrated class score is $S_c=P_c^{base}w_c=(N_c/N)w_c$. To preserve intra-class fairness, every instance in class $c$ shares this score equally. Thus an instance $x_i$ with $y_i=c$ has unnormalized score $s_{x_i}=S_c/N_c=w_c/N$. The final probability is
\begin{equation}\label{equ:instance_prob}
\begin{aligned}
P_{x_i}
&=
\frac{s_{x_i}}{\sum_{j=1}^{N}s_{x_j}}
=
\frac{w_{y_i}}{N\sum_{c=1}^{C}P_c^{base}w_c} .
\end{aligned}
\end{equation}
Equation~\eqref{equ:instance_prob} specifies how the dynamic class weight is translated into the target sampling distribution. SGS reweights the long-tailed prior by the frequency-sharpness score. The frequency term raises $q_c(t)$ for classes that have been sampled less often, while the sharpness term lowers $q_c(t)$ for classes whose perturbation response is currently large, preventing the next mini-batches from being dominated by unstable high-sharpness updates. After each mini-batch, $m_c(t)$ and $h_c(t)$ are updated, so the sampler forms a closed-loop control: under-sampled classes are pulled back into the batch stream, whereas classes with excessive perturbation sensitivity are temporarily relaxed. The factor $1/N_c$ assigns this class-level probability uniformly to instances within the class, preserving intra-class fairness.

\begin{algorithm}[t]
\caption{SGS Training}
\label{alg:sges}
\begin{algorithmic}[1]
\REQUIRE Training set $\mathcal{D}$, model $f_{\theta}$, epochs $T$, batch size $B$, warm-up ratio $r_w$, mixing schedule $\alpha(e)$, maximum ratio $\alpha_{\max}$, SAM radius $\rho$, EMA momentum $\beta$, sharpness exponent $\kappa$, constant $\xi$
\ENSURE Trained model parameters $\theta_T$
\STATE Initialize $m_c\leftarrow 1$, $h_c\leftarrow 1$, $P_{x_i}\leftarrow P^{base}_{x_i}$, and $E_w=\lfloor r_wT\rfloor$
\FOR{$e=1$ to $T$}
\STATE Set $\alpha(e)$ according to the warm-up schedule
\FOR{each mini-batch}
\STATE By Eq.~\eqref{equ:mixed_batch_counts}, draw $B_{\mathrm{base}}(e)$ samples from $P^{base}_{x_i}$ and $B_{\mathrm{SGS}}(e)$ samples from $P_{x_i}$ to form $B_t$
\STATE Perform the SAM ascent--descent update and record $\Delta L_{x_i}=|\ell_i(\theta+\epsilon_t)-\ell_i(\theta)|$
\STATE Update class counts $m_c\leftarrow m_c+|B_{t,c}|$
\STATE If $e>E_w$, update $h_c\leftarrow\beta h_c+(1-\beta)|B_{t,c}|^{-1}\sum_{x_i\in B_{t,c}}\Delta L_{x_i}$
\STATE If $e>E_w$, update $P_{x_i}$ by Eqs.~\eqref{equ:class_w_final} and \eqref{equ:instance_prob}
\ENDFOR
\ENDFOR
\STATE \textbf{return} $\theta_T$
\end{algorithmic}
\end{algorithm}

To maintain stable training, SGS constructs each mini-batch from the original long-tailed distribution $P^{base}_{x_i}$ and the current SGS distribution $P_{x_i}$ defined in Eq.~\eqref{equ:instance_prob}. We use a warm-up ratio $r_w\in[0,1]$ and set $E_w=\lfloor r_wT\rfloor$. For a mini-batch of size $B$, the numbers assigned to the two distributions at epoch $e$ are
\begin{equation}\label{equ:mixed_batch_counts}
\begin{aligned}
B_{\mathrm{base}}(e)
&=\operatorname{round}\!\left((1-\alpha(e))B\right),\\
B_{\mathrm{SGS}}(e)
&=B-B_{\mathrm{base}}(e)
\end{aligned}
\end{equation}
respectively. We first draw $B_{\mathrm{base}}(e)$ samples from $P^{base}_{x_i}$, then draw $B_{\mathrm{SGS}}(e)$ samples from $P_{x_i}$ after excluding the selected indices, and concatenate the two subsets into a mini-batch of size $B$. During warm-up, $\alpha(e)=0$, so the entire mini-batch follows the original sampler. After warm-up, $\alpha(e)$ gradually increases toward $\alpha_{\max}$, increasing the SGS portion without abruptly changing the training distribution. Unless otherwise specified, we set $\alpha_{\max}=1$. Algorithm~\ref{alg:sges} summarizes the procedure.

\subsection{A Stochastic Dynamics Perspective of SGS}

We further interpret SGS as a continuous distributional dynamics. This view characterizes how frequency-curvature feedback moves the empirical long-tailed distribution toward a flatness-aware target distribution.

\begin{theorem}[Continuous-Time Dynamics of SGS]\label{thm:sde_dynamics}
Let $q_i(t)$ be the instance-level sampling probability and $q_c(t)=\sum_{i:y_i=c}q_i(t)$ be the class-level probability. Define the frequency-curvature drive
$g_c(t)=\log(p_c(t)+\xi)+\log(\tilde{h}_c(t)^{\kappa}+\xi)$,
so the class score is proportional to $\exp(-g_c(t))$. Assume $g_c(t)$ admits the Stratonovich limit
$dg_c(t)=\mu_c(t)dt+\sigma_c(t)\circ dW_c(t)$.
Then, for an instance $z_i$ with label $y_i$, the induced sampling probability satisfies
\begin{equation}\label{equ:sde_dynamics}
\begin{aligned}
&dq_i(t)
= - q_i(t) \left[ \mu_{y_i}(t) - \bar{\mu}(t) \right] dt\\
&\quad - q_i(t)\Big[
\sigma_{y_i}(t)\circ dW_{y_i}(t)
- \sum_{c=1}^C q_c(t)\sigma_c(t)\circ dW_c(t)
\Big].
\end{aligned}
\end{equation} 
where $\bar{\mu}(t)=\sum_{c=1}^C q_c(t)\mu_c(t)$.
\end{theorem}

\textbf{Remark.} This SDE explicitly decouples the sampling evolution into a deterministic relative competition (the drift term) and a stochastic fluctuation (the diffusion term). The drift establishes a self-regulating mechanism: probability mass intrinsically flows to classes whose evolution rate $\mu_{y_i}(t)$ is below the global baseline $\bar{\mu}(t)$. This ensures a continuous transfer of sampling weights from over-represented head classes to tail classes. Crucially, the curvature component acts as a geometry-aware "thermostat"—if a tail class enters an excessively sharp minimum, its curvature penalty spikes, triggering immediate negative feedback to suppress its sampling rate. Simultaneously, the diffusion term formally models the destructive variance caused by sparse tail-class mini-batch sampling. This motivates the use of EMA to smooth class-level estimates and obtain a more stable curvature-aware sampling distribution.

\begin{theorem}[Stationary Density for a Fixed Class]\label{thm:sde_stationary}
Consider the sampling probability of a fixed class $c$, denoted by $q=q_c(t)$. In a local time window, freeze the slowly varying coefficients and write $a_c=\mu_c-\bar{\mu}$, where $\bar{\mu}=\sum_{j=1}^C q_j\mu_j$. Assume the remaining class-wise noise can be represented by an effective volatility $\tilde{\sigma}_c$ and the one-dimensional dynamics are
\begin{equation}
dq=-a_cq\,dt+q\tilde{\sigma}_c\circ dW_t .
\end{equation}
On a compact positive-probability support $q\in[q_{\min},q_{\max}]\subset(0,1)$ with zero probability flux at the boundaries, the stationary density satisfies
\begin{equation}\label{equ:sde_stationary}
\varphi_{\infty}^{(c)}(q)=\frac{1}{z_c}q^{-\gamma_c}, \qquad
\gamma_c
=1+\frac{2a_c}{\tilde{\sigma}_c^2}
=1+\frac{2(\mu_c-\bar{\mu})}{\tilde{\sigma}_c^2},
\end{equation}
where $z_c=\int_{q_{\min}}^{q_{\max}}q^{-\gamma_c}dq$ normalizes the density.
\end{theorem}

\textbf{Remark.}
The equilibrium is governed by the relative drift $a_c=\mu_c-\bar{\mu}$.
When $\mu_c<\bar{\mu}$, we have $\gamma_c<1$, so the stationary density is less concentrated near the lower-probability boundary and the class tends to retain larger exposure. When $\mu_c>\bar{\mu}$, $\gamma_c$ increases and the density shifts toward smaller sampling probabilities, suppressing over-sampled or sharper classes. Since $\varphi_{\infty}^{(c)}(q)\propto \exp(-\gamma_c\log q)$, changes in the frequency-curvature drive affect the sampling density in a nonlinear way. Thus SGS does not behave as naive tail over-sampling: it favors classes with insufficient exposure or moderate curvature, while reducing the probability of classes whose frequency-curvature drive is already high. A balanced fixed point corresponds to aligning the class-wise drive with the population average, yielding a frequency-curvature balanced sampling target.

\begin{theorem}[Tighter PAC-Bayes Bound under SGS Sampling]\label{thm:diff_uperbound}
Let $\mathbf{u}$ be the class-balanced prior, $\boldsymbol{\pi}$ be the empirical long-tailed prior, and $\mathbf{q}^{\star}$ be the stationary SGS distribution from \cref{thm:sde_stationary}. Let $\widehat{\mathcal{L}}_{\mathbf{q},S}^{\rho}(w)$ denote the SAM sharp empirical risk under class-level distribution $\mathbf{q}$. We write $\mathcal{B}(\mathbf{q})$ for the sampling-dependent PAC penalty, consisting of the balanced-prior mismatch and the class-wise effective-sample complexity:
\begin{equation}
\mathcal{B}(\mathbf{q})
=
\|\mathbf{q}-\mathbf{u}\|_1
+
\frac{1}{C}\sum_{c=1}^{C}\Phi(q_c),
\end{equation}
where $\Phi(q_c)$ denotes the PAC complexity term for class $c$ and decreases as the effective class sample size $nq_c$ increases. Assume the sharp loss is bounded and each class has positive effective sample size. Suppose that the movement from $\boldsymbol{\pi}$ to $\mathbf{q}^{\star}$ admits an equalizing-transfer decomposition in the effective-sample region, that $\mathbf{q}^{\star}$ is closer to the balanced prior,
\begin{equation}
\|\mathbf{q}^{\star}-\mathbf{u}\|_1
<
\|\boldsymbol{\pi}-\mathbf{u}\|_1,
\end{equation}
and that any increase in sharp empirical risk is smaller than the resulting reduction in $\mathcal{B}$. Then, with probability at least $1-\delta$, SGS has a tighter PAC-Bayes right-hand side than ordinary long-tailed SAM:
\begin{equation}\label{equ:sges_tighter_than_ltsam_main}
\widehat{\mathcal{L}}_{\mathbf{q}^{\star},S}^{\rho}(w)
+
\mathcal{B}(\mathbf{q}^{\star})
<
\widehat{\mathcal{L}}_{\boldsymbol{\pi},S}^{\rho}(w)
+
\mathcal{B}(\boldsymbol{\pi}) .
\end{equation}
\end{theorem}

\textbf{Remark.}
Compared with the standard SAM PAC-Bayes analysis under the empirical training distribution, ordinary SAM on long-tailed data suffers from two coupled effects. First, the empirical SAM objective is optimized under the imbalanced prior $\boldsymbol{\pi}$, whereas long-tailed recognition is evaluated by the class-balanced risk, yielding the mismatch term $\|\boldsymbol{\pi}-\mathbf{u}\|_1$. Second, tail classes have $n\pi_c\ll n/C$, so their class-wise PAC complexity scales roughly as $\mathcal{O}((n\pi_c)^{-1/2})$ rather than the balanced $\mathcal{O}((n/C)^{-1/2})$. Thus, even if SAM encourages flat minima locally, its bound can remain loose because rare classes have insufficient effective samples. The stationary law in \cref{thm:sde_stationary} explains how SGS reduces this looseness: it shifts probability away from over-exposed or high-curvature classes and toward classes with lower frequency-curvature drive. As a result, $\|\mathbf{q}^{\star}-\mathbf{u}\|_1$ is reduced, the effective denominators $nQ_c^{\star}-1$ for tail classes become larger, and the sharp empirical term is less dominated by over-represented sharp regions.

\section{Experiments}
\subsection{Hyperparameter Analysis}
\label{subsec:hyperparameter_analysis}

\noindent\textbf{Warm-Up Ratio $r_w$.} Table~\ref{tab:warmup_analysis} studies the warm-up ratio $r_w$, which controls when SGES starts to use frequency-curvature feedback. A too large $r_w$ delays geometry-aware sampling and leaves insufficient time for tail correction. For example, on CIFAR-100 LT with IR100, $r_w=0.9$ achieves the best head accuracy but only obtains 12.83 tail accuracy and 44.94 overall accuracy. In contrast, activating SGES too early can be unstable because early sharpness estimates are noisy. The best IR100 performance is achieved at $r_w=0.7$, improving tail accuracy to 20.76 and overall accuracy to 47.06. Under the more severe IR200 setting, the best result appears at $r_w=0.4$, indicating that stronger imbalance requires earlier intervention. Overall, a moderate warm-up is important: it preserves stable representation learning at the beginning while leaving enough epochs for SGES to reshape the sampling trajectory toward tail- and curvature-balanced optimization. In practice, we use a larger $r_w$ for milder imbalance and a smaller $r_w$ for more severe imbalance.

\begin{table}[t]
  \centering
  \caption{Effect of the warm-up ratio $r_w$ on CIFAR-100 LT. IR100 reports Head, Medium, Tail, and All accuracy; IR200 reports overall accuracy.}
  \begin{tabular}{c|cccc|c}
    \toprule
    \multirow{2}{*}{$r_w$} & \multicolumn{4}{c|}{IR100} & IR200 \\
          & Head & Med. & Tail & All & All \\
    \midrule
    0.9 & \textbf{73.47} & 44.59 & 12.83 & 44.94 & 39.47 \\
    0.8 & 71.88 & 45.06 & 19.31 & 46.75 & 42.37 \\
    0.7 & 71.91 & 44.82 & \textbf{20.76} & \textbf{47.06} & 41.67 \\
    0.6 & 73.06 & 44.82 & 19.69 & 46.86 & 42.31 \\
    0.5 & 70.97 & 43.50 & 20.38 & 46.03 & 42.13 \\
    0.4 & 70.88 & 44.56 & 20.07 & 46.38 & \textbf{42.84} \\
    0.3 & 69.71 & \textbf{45.15} & 19.66 & 46.17 & 41.75 \\
    0.2 & 69.94 & 43.68 & 19.52 & 45.52 & 41.70 \\
    0.1 & 70.03 & 43.91 & 18.79 & 45.42 & 41.28 \\
    \bottomrule
  \end{tabular}
  \label{tab:warmup_analysis}
\end{table}

\noindent\textbf{Sharpness Exponent $\kappa$.} Table~\ref{tab:kappa_analysis} studies the sharpness exponent $\kappa$, which controls the strength of the curvature term in the sampling score. On CIFAR-100 LT with IR100, decreasing $\kappa$ from 1.0 to 0.6 improves the overall accuracy from 47.56 to 48.19 and gives the best medium-class accuracy. This indicates that smoothing the curvature signal with $\kappa=0.6$ improves the balance between representation stability and tail exposure. In contrast, $\kappa=0.9$ gives the best tail accuracy but lower overall accuracy, while $\kappa=0.1$ and $\kappa=0.3$ weaken the sharpness signal and reduce tail performance. For the milder IR10 setting, performance is less sensitive to $\kappa$. Based on this analysis, we set $\kappa=0.6$ unless otherwise specified.

\begin{table}[t]
  \centering
  \caption{Effect of the sharpness exponent $\kappa$ on CIFAR-100 LT. IR100 reports Head, Medium, Tail, and All accuracy; IR10 reports overall accuracy.}
  \begin{tabular}{c|cccc|c}
    \toprule
    \multirow{2}{*}{$\kappa$} & \multicolumn{4}{c|}{IR100} & IR10 \\
          & Head & Med. & Tail & All & All \\
    \midrule
    1.0 & \textbf{72.65} & 44.18 & 20.62 & 46.94 & 61.62 \\
    0.9 & 70.91 & 44.82 & \textbf{23.62} & 47.52 & 62.43 \\
    0.8 & 71.79 & 46.41 & 22.41 & 47.87 & 62.33 \\
    0.7 & 71.32 & 46.85 & 21.17 & 47.45 & 62.41 \\
    0.6 & 71.24 & \textbf{48.88} & 20.83 & \textbf{48.19} & 63.03 \\
    0.5 & 70.15 & 48.41 & 16.14 & 46.31 & \textbf{63.46} \\
    0.4 & 71.47 & 47.97 & 16.97 & 46.62 & 62.72 \\
    0.3 & 71.50 & 47.41 & 15.35 & 46.15 & 62.79 \\
    0.2 & 72.32 & 47.65 & 17.76 & 47.13 & 62.64 \\
    0.1 & 72.47 & 46.00 & 13.90 & 45.42 & 62.49 \\
    \bottomrule
  \end{tabular}
  \label{tab:kappa_analysis}
\end{table}

\begin{table*}[ht]
  \centering
  \caption{Performance comparison on CIFAR-LT datasets with various imbalance ratios (IR). For IR100, we report accuracy on Head, Medium, Tail, and All classes; for other IRs, we report overall accuracy.}
  \renewcommand{\arraystretch}{1.08}
    \begin{tabular}{@{}cl|cccc|ccc@{}}
    \toprule
    \multirow{2}{*}{Dataset} & \multirow{2}{*}{Method} & \multicolumn{4}{c|}{IR100} & IR200 & IR50 & IR10 \\
          &        & Head  & Med   & Tail  & All   & All   & All  & All \\
    \midrule
    \multirow{12}{*}{\rotatebox[origin=c]{90}{CIFAR-10 LT}}
    & CE    & 87.0  & 73.6  & 54.0  & 73.1  & 68.6  & 78.3  & 87.4 \\
    & CE+SAM & 89.5  & 73.9  & 56.7  & 75.0  & 69.8  & 79.6  & 88.8 \\
    & CE+ImbSAM & 88.0  & \textbf{79.0}  & 60.1  & 76.9  & 72.6  & 81.1  & 89.3 \\
    & CE+CC-SAM & 88.9  & 74.1  & 61.3  & 76.2  & 71.3  & 80.0  & 89.2 \\
    & CE+Focal-SAM & 89.3  & 75.4  & 62.9  & 77.2  & 71.7  & 82.0  & \textbf{90.0} \\
    & \textbf{CE+SGS-SAM} & \textbf{93.93{\scriptsize$\pm$0.47}} & 76.62{\scriptsize$\pm$0.85} & \textbf{64.70{\scriptsize$\pm$1.78}} & \textbf{77.96{\scriptsize$\pm$0.15}} & \textbf{73.53{\scriptsize$\pm$0.67}} & \textbf{82.39{\scriptsize$\pm$0.10}} & 89.87{\scriptsize$\pm$0.21} \\
    \cmidrule(lr){2-9}
    & LA & \textbf{87.6} & 72.6  & 70.1  & 77.9  & 74.3  & 81.6  & 87.8 \\
    & LA+SAM & 86.7  & 80.6  & 78.2  & 82.3  & 78.9  & 85.4  & 90.2 \\
    & LA+ImbSAM & 84.1  & 81.6  & 80.1  & 82.2  & 78.6  & 84.7  & 89.5 \\
    & LA+CC-SAM & 86.6  & 80.8  & 78.5  & 82.5  & 79.1  & 85.5  & 90.2 \\
    & LA+Focal-SAM & 86.9  & 81.2  & 79.2  & 82.9  & 79.6  & 85.5  & \textbf{90.5} \\
    & \textbf{LA+SGS-SAM} & 83.97{\scriptsize$\pm$1.07} & \textbf{83.29{\scriptsize$\pm$0.31}} & \textbf{84.96{\scriptsize$\pm$0.71}} & \textbf{84.06{\scriptsize$\pm$0.28}} & \textbf{81.65{\scriptsize$\pm$0.14}} & \textbf{86.56{\scriptsize$\pm$0.19}} & 90.39{\scriptsize$\pm$0.18} \\
    \midrule
    \multirow{12}{*}{\rotatebox[origin=c]{90}{CIFAR-100 LT}}
    & CE    & 69.2  & 41.6  & 9.0   & 41.5  & 37.5  & 45.6  & 58.1 \\
    & CE+SAM & 72.7  & 41.8  & 7.0   & 42.2  & 38.9  & 46.8  & 59.7 \\
    & CE+ImbSAM & 68.5  & 46.0  & 9.6   & 43.0  & 38.7  & 47.8  & 60.1 \\
    & CE+CC-SAM & 70.1  & 44.2  & 9.0   & 42.7  & 39.1  & 47.4  & 60.0 \\
    & CE+Focal-SAM & \textbf{73.8} & 44.2  & 8.9   & 44.0  & 39.6  & 48.1  & 60.9 \\
    & \textbf{CE+SGS-SAM} & 72.97{\scriptsize$\pm$0.40} & \textbf{46.58{\scriptsize$\pm$1.57}} & \textbf{19.75{\scriptsize$\pm$0.57}} & \textbf{47.56{\scriptsize$\pm$0.79}} & \textbf{42.99{\scriptsize$\pm$0.61}} & \textbf{52.46{\scriptsize$\pm$0.16}} & \textbf{62.77{\scriptsize$\pm$0.31}} \\
    \cmidrule(lr){2-9}
    & LA & 61.3  & 42.3  & 28.6  & 44.9  & 41.8  & 50.3  & 59.4 \\
    & LA+SAM & 63.1  & 52.2  & 32.2  & 50.0  & 45.5  & 52.8  & 62.6 \\
    & LA+ImbSAM & 57.4  & 51.1  & 31.0  & 47.3  & 43.4  & 52.2  & 62.4 \\
    & LA+CC-SAM & 63.7  & 51.9  & 32.3  & 50.1  & 45.6  & 53.0  & 63.0 \\
    & LA+Focal-SAM & \textbf{63.9} & 53.0  & 32.5  & 50.7  & 46.0  & 54.5  & \textbf{63.8} \\
    & \textbf{LA+SGS-SAM} & 62.24{\scriptsize$\pm$0.53} & \textbf{53.53{\scriptsize$\pm$0.26}} & \textbf{38.22{\scriptsize$\pm$0.28}} & \textbf{51.99{\scriptsize$\pm$0.28}} & \textbf{46.69{\scriptsize$\pm$0.27}} & \textbf{56.05{\scriptsize$\pm$0.36}} & 63.16{\scriptsize$\pm$0.27} \\
    \bottomrule
    \end{tabular}%
  \label{tab: performance comparison on CIFAR-LT with imbalance ratio of 100}%
\end{table*}%

\subsection{Long-Tailed Classification}
\label{subsec: experiment protocols}

\textbf{Datasets \& Evaluation.} We evaluate our method on four standard long-tailed (LT) recognition benchmarks: CIFAR-10 LT, CIFAR-100 LT~\cite{cao2019learning} (with imbalance ratios IR in $\{10, 50, 100, 200\}$), ImageNet-LT~\cite{liu2019large}. CIFAR100-LT and ImageNet-LT are artificially truncated from the balanced CIFAR100~\cite{krizhevsky2009learning} and ImageNet~\cite{liu2019large} datasets. Performance is measured using top-1 balanced accuracy across the overall test set, as well as on three disjoint subset splits: Head, Medium, and Tail. 

\textbf{Architectures \& Training.} We employ ResNet-32~\cite{he2016deep} for the CIFAR datasets and ResNet-50~\cite{he2016deep} for ImageNet-LT and iNaturalist, optimizing all models for 200 epochs. We set SGD optimizer with momentum 0.9 as the base optimizer and train all models for 200 epochs, with a batch size of 64 for CIFAR100-LT and 256 for ImageNet-LT. Besides the Cross-Entropy loss (CE) setting, we also evaluate SGS with Logit Adjustment loss (LA)~\cite{menonlong}. LA adds the empirical class-prior offset to the logits and can be applied either as a post-hoc correction or as a training loss for balanced error. However, SGS itself changes the mini-batch sampling distribution, so directly applying LA throughout SGS training would couple two class-prior corrections. We therefore adopt a deferred LA protocol: first train SGS-SAM for 180 epochs to learn the representation and geometry-aware sampling trajectory, then freeze the backbone and fine-tune only the classifier for 20 epochs with the LA loss and SAM under the original long-tailed training set. More details about datasets, networks, and training settings are presented in Appendix.

\textbf{Compared Methods.} We compare against a broad spectrum of state-of-the-art approaches. On ResNet architectures, we integrate standard LT loss functions with vanilla SAM~\cite{foretsharpness}, ImbSAM~\cite{zhou2023imbsam}, and CC-SAM~\cite{zhou2023class}, Focal-SAM~\cite{li2025focal} alongside other representative LT baselines.

\noindent\textbf{Comparison on CIFAR10/100-LT.}
Table~\ref{tab: performance comparison on CIFAR-LT with imbalance ratio of 100} summarizes the CIFAR-LT results under different imbalance ratios. SGS-SAM consistently improves overall accuracy in high-imbalance regimes. For IR100, it improves CIFAR-10 LT from 77.2 to 77.96 under CE and from 82.9 to 84.06 under LA; on CIFAR-100 LT, it improves from 44.0 to 47.56 under CE and from 50.7 to 51.99 under LA. Similar gains are observed for IR200 and IR50, where SGS-SAM achieves the best overall accuracy across both datasets and loss settings.

The gains mainly come from medium and tail classes, which are typically under-trained in long-tailed recognition. On CIFAR-10 LT with IR100, CE+SGS-SAM improves tail accuracy over CE+Focal-SAM by 1.80 points, and LA+SGS-SAM improves medium/tail accuracy from 81.2/79.2 to 83.29/84.96. On CIFAR-100 LT, CE+SGS-SAM raises tail accuracy from 8.9 to 19.75, while LA+SGS-SAM raises it from 32.5 to 38.22. These results support the motivation of SGS: reallocating training exposure by both class frequency and local sharpness gives under-represented classes more effective optimization opportunities.

SGS also remains compatible with deferred logit-adjusted calibration. LA adjusts the decision boundary through class-prior logits, while SGS shapes the training trajectory through geometry-aware exposure; applying LA only in the final fine-tuning stage avoids disturbing the SGS sampling dynamics. When the imbalance becomes mild (IR10), the advantage of SGS-SAM becomes smaller and is generally comparable to Focal-SAM, indicating that geometry-aware sampling is most useful when the training distribution is severely skewed.

\begin{table}[t]
  \centering
  \caption{Performance comparison on ImageNet-LT. We report top-1 accuracy on Head, Medium, Tail, and All classes. The CE block follows ImbSAM~\cite{zhou2023imbsam}, and the LA block follows Focal-SAM~\cite{li2025focal}.}
  \setlength{\tabcolsep}{1mm}
  \begin{tabular}{@{}lcccc@{}}
  \toprule
  Method & Head & Med & Tail & All \\
  \midrule
  CE & 69.3 & 41.7 & 10.3 & 48.2 \\
  CE+SAM & \textbf{70.0} & 41.1 & 10.2 & 48.2 \\
  CE+ImbSAM & 68.5 & 47.5 & 21.6 & 52.2 \\
  \textbf{CE+SGS-SAM} & 66.33{\tiny$\pm$0.05} & \textbf{50.17{\tiny$\pm$0.16}} & \textbf{28.19{\tiny$\pm$0.12}} & \textbf{53.40{\tiny$\pm$0.11}} \\
  \midrule
  LA & 62.8 & 49.0 & 31.8 & 52.0 \\
  LA+SAM & 63.1 & 51.7 & 33.1 & 53.6 \\
  LA+ImbSAM & 62.6 & 50.3 & 32.6 & 52.6 \\
  LA+Focal-SAM & \textbf{63.9} & 52.2 & 34.4 & 54.3 \\
  \textbf{LA+SGS-SAM} & 63.11{\tiny$\pm$0.90} & \textbf{52.90{\tiny$\pm$0.20}} & \textbf{36.26{\tiny$\pm$2.83}} & \textbf{54.57{\tiny$\pm$0.05}} \\
  \bottomrule
  \end{tabular}
  \label{tab:imagenet_lt_results}
\end{table}

\begin{table}[t]
  \centering
  \caption{Foundation-model fine-tuning results on ImageNet-LT and
  iNaturalist. Results of compared methods are reported by
  Focal-SAM~\cite{li2025focal}. We report top-1 accuracy.}
  \setlength{\tabcolsep}{0.8mm}
  \begin{tabular}{@{}lcccccccc@{}}
    \toprule
    \multirow{2}{*}{Method}
    & \multicolumn{4}{c}{ImageNet-LT}
    & \multicolumn{4}{c}{iNaturalist} \\
    \cmidrule(lr){2-5}\cmidrule(lr){6-9}
    & Head & Med & Tail & All
    & Head & Med & Tail & All \\
    \midrule

    \multicolumn{9}{@{}l}{\textit{Full fine-tuning (FFT)}} \\
    Base
    & 79.9 & 70.5 & 51.0 & 71.5
    & \textbf{69.7} & 71.9 & 71.7 & 71.6 \\

    SAM
    & \textbf{80.9} & 72.9 & 54.3 & 73.5
    & 69.5 & \textbf{74.4} & 74.4 & 73.8 \\

    ImbSAM
    & 80.6 & 72.6 & 52.2 & 72.9
    & 68.5 & 73.4 & 73.8 & 73.1 \\

    CC-SAM
    & 80.6 & 73.6 & 54.2 & 73.6
    & 69.2 & 74.1 & 74.2 & 73.6 \\

    Focal-SAM
    & 80.8 & 73.9 & 54.4 & 73.9
    & 69.1 & 74.7 & 74.3 & 74.0 \\

    \textbf{SGS-SAM}
    & 80.0 & \textbf{74.5} & \textbf{64.9} & \textbf{75.3}
    & 69.2 & 73.8 & 74.7 & 73.7 \\

    \textbf{SGS-SAM$_{25}$}
    & -- & -- & -- & --
    & 69.6 & 74.2 & \textbf{75.1} & \textbf{74.1} \\

    \midrule

    \multicolumn{9}{@{}l}{\textit{LIFT fine-tuning (LIFT)}} \\
    Base
    & 79.7 & 76.2 & 72.8 & 77.1
    & 74.1 & 79.4 & 81.5 & 79.7 \\

    SAM
    & \textbf{79.9} & 76.4 & 72.7 & 77.2
    & 73.5 & 79.7 & 81.6 & 79.8 \\

    ImbSAM
    & 79.8 & 76.4 & 72.5 & 77.2
    & 73.2 & 79.5 & 81.4 & 79.6 \\

    CC-SAM
    & 79.8 & 76.4 & 73.3 & 77.3
    & \textbf{74.0} & 79.4 & 81.5 & 79.7 \\

    Focal-SAM
    & 79.7 & \textbf{76.6} & \textbf{73.6} & 77.4
    & 73.9 & 79.8 & 81.7 & 80.0 \\

    \textbf{SGS-SAM}
    & 79.7 & \textbf{76.6} & 73.4 & \textbf{77.4}
    & 72.9 & 79.6 & 81.6 & 79.7 \\

    \textbf{SGS-SAM$_{25}$}
    & -- & -- & -- & --
    & \textbf{74.0} & \textbf{80.2} & \textbf{82.1} & \textbf{80.3} \\

    \bottomrule
  \end{tabular}
  \label{tab:large_scale_finetuning}
\end{table}

\noindent\textbf{ImageNet-LT Results.}
Table~\ref{tab:imagenet_lt_results} further evaluates SGS-SAM on the large-scale ImageNet-LT benchmark. Under CE training, vanilla SAM does not improve the overall accuracy over CE and leaves tail accuracy almost unchanged, suggesting that class-agnostic flatness optimization is still dominated by head-class exposure. ImbSAM improves tail accuracy to 21.6 by modifying the perturbation strategy, while CE+SGS-SAM further increases tail accuracy to 28.19 and overall accuracy to 53.40. Under the deferred LA protocol, LA+SGS-SAM also improves over LA+Focal-SAM, raising tail accuracy from 34.4 to 36.26 and overall accuracy from 54.3 to 54.57. The head accuracy is slightly lower because SGS deliberately reallocates training exposure from head-dominated batches toward medium and tail classes. These results are consistent with the CIFAR-LT observations and show that geometry-aware sampling remains effective on a larger-scale long-tailed benchmark.

\subsection{Large-Scale Long-Tailed Fine-Tuning}
\label{subsec:large_scale_finetuning}

\textbf{Fine-Tuning Protocol.}
We further evaluate SGS for foundation-model fine-tuning on ImageNet-LT and iNaturalist~\cite{van2018inaturalist}, following the experimental protocols of Focal-SAM~\cite{li2025focal} and LIFT~\cite{shi2024long}. Specifically, we fine-tune the CLIP~\cite{radford2021learning} image encoder with a ViT-B/16 backbone~\cite{dosovitskiyimage} under two fine-tuning paradigms, namely full fine-tuning (FFT) and LIFT. The default training schedule consists of 20 epochs. Since this setting employs the LA loss defined according to the original long-tailed class distribution, we set the maximum SGS mixing coefficient to $\alpha_{\max}=0.1$ to prevent an excessive shift in the effective sampling distribution. Considering the short fine-tuning schedule, we set $r_w=0$ to activate SGS from the beginning and use $\kappa=2$ to strengthen the sharpness feedback. On iNaturalist, the SGS variants continue to improve beyond the default schedule. We therefore additionally report their results after 25 epochs, denoted by SGS-SAM$_{25}$ in Table~\ref{tab:large_scale_finetuning}.

\noindent\textbf{Fine-Tuning Results.}
Table~\ref{tab:large_scale_finetuning} summarizes the results under the CLIP/ViT-B/16 fine-tuning protocol. Under FFT on ImageNet-LT, SGS-SAM achieves the best medium-, tail-, and overall accuracies of 74.5\%, 64.9\%, and 75.3\%, respectively. Compared with Focal-SAM, it improves tail accuracy by 10.5 percentage points and overall accuracy by 1.4 points, showing that geometry-aware sampling substantially benefits underrepresented classes. Under LIFT, SGS-SAM achieves 77.4\% overall accuracy, matching the best result in this group, while obtaining the best medium-class accuracy of 76.6\% and maintaining competitive head- and tail-class performance.

On iNaturalist, the 20-epoch SGS-SAM variants remain competitive but have not fully converged. Extending FFT with SGS-SAM to 25 epochs increases tail and overall accuracy to 75.1\% and 74.1\%, respectively, outperforming the corresponding 20-epoch result and yielding the best tail accuracy within the FFT group. More substantial gains are observed under LIFT: LIFT with SGS-SAM$*{25}$ achieves head, medium, tail, and overall accuracies of 74.0\%, 80.2\%, 82.1\%, and 80.3\%, respectively. It obtains the best medium-, tail-, and overall results in the LIFT group and ties the best head-class accuracy. Although SGS-SAM$*{25}$ uses five additional epochs, its total training time is only $1.26\times$ that of vanilla SAM, remaining below the $1.29\times$ cost of Focal-SAM (the reasons why SGS needs more epoch for better results is discussed in Appendix). These results demonstrate that SGS can improve long-tailed foundation-model fine-tuning while preserving favorable training efficiency.

\subsection{Training Speed}
To reduce the influence of hardware and implementation details, we report relative training time with respect to vanilla SAM instead of absolute seconds. The comparison ratios in Table~\ref{tab:training_speed} are taken from the Focal-SAM~\cite{li2025focal}.

SGS-SAM follows the same optimization pipeline as vanilla SAM. The sharpness signal is directly obtained from the loss gap produced by the original SAM ascent--descent steps, while the sampler only maintains class-level frequency and EMA sharpness statistics. Therefore, SGS does not introduce additional class-wise perturbations or extra backward passes. Its overhead is limited to updating a (C)-dimensional sampling distribution and drawing mini-batches accordingly, which is negligible compared with the forward/backward computation of deep networks.

\begin{table}[t]
  \centering
  \caption{Relative training time per epoch on CIFAR-LT datasets. Ratios are measured with respect to vanilla SAM.}
  \begin{tabular}{lcc}
  \toprule
  Method & CIFAR-10 LT & CIFAR-100 LT \\
  \midrule
  SAM & 1.00$\times$ & 1.00$\times$ \\
  ImbSAM & 1.37$\times$ & 1.39$\times$ \\
  CC-SAM & 2.05$\times$ & 4.10$\times$ \\
  Focal-SAM & 1.36$\times$ & 1.40$\times$ \\
  \textbf{SGS-SAM} & \textbf{1.02$\times$} & \textbf{1.02$\times$} \\
  \bottomrule
  \end{tabular}
  \label{tab:training_speed}
\end{table}

\section{Conclusion}
In this paper, we presented Sharpness-Guided Equilibrium Sampling (SGES), a geometry-aware sampling framework for long-tailed recognition. Different from existing SAM-based methods that mainly modify the loss or perturbation rule after biased mini-batches are drawn, SGES directly reshapes the training trajectory by coupling class-frequency feedback with local sharpness estimates. The resulting sampler assigns more effective optimization opportunities to under-represented classes while avoiding excessive updates from unstable high-sharpness regions. We further introduced a warm-up and progressive mixing strategy to stabilize early training, and provided a stochastic-dynamics view that explains how the discrete sampler evolves toward a frequency-curvature balanced distribution. Extensive experiments on CIFAR-LT and ImageNet-LT show that SGES-SAM consistently improves medium and tail performance under severe imbalance, and remains compatible with deferred logit-adjusted calibration. We also extend the protocol to long-tailed instance segmentation through decoupled classifier training on LVIS. Since SGES reuses the sharpness information already produced by SAM and only updates lightweight class-level statistics, it offers an efficient alternative to class-wise perturbation methods. A limitation is that the sampler depends on reliable class-level sharpness statistics, so the warm-up ratio and sharpness exponent still require mild tuning under different imbalance ratios. Overall, our results suggest that controlling the sampling process with optimization geometry is a promising direction for scalable long-tailed learning.

\bibliography{aaai2026}


\end{document}


\makeatletter
\let\AAAIappendix@twocolumn\twocolumn
\renewcommand{\twocolumn}[1][]{#1}
\onecolumn
\maketitle
\let\twocolumn\AAAIappendix@twocolumn
\makeatother

\section{Proofs}
\subsection{Proof of Theorem 1}
\begin{proof}
We start from the class-level sampling score used by SGS. For class $c$, define the frequency-curvature drive
\begin{equation}
g_c(t)=\log(p_c(t)+\xi)+\log(\tilde{h}_c(t)^{\kappa}+\xi),
\end{equation}
and let $\tilde{w}_c(t)=\exp(-g_c(t))$. The exponent $\kappa$ controls the strength of the sharpness feedback while preserving the monotonicity of the curvature term. The mean normalization used in the algorithm is a global scalar across classes and therefore cancels after the class probabilities are normalized. Thus the class-level sampling distribution can be written as
\begin{equation}
q_c(t)=
\frac{p_c^{\mathrm{base}}\tilde{w}_c(t)}
{\sum_{j=1}^C p_j^{\mathrm{base}}\tilde{w}_j(t)}
=
\frac{p_c^{\mathrm{base}}\exp(-g_c(t))}
{\sum_{j=1}^C p_j^{\mathrm{base}}\exp(-g_j(t))}.
\end{equation}
For notational convenience, define
\begin{equation}
r_c(t)=p_c^{\mathrm{base}}\tilde{w}_c(t),\qquad
z(t)=\sum_{j=1}^C r_j(t),
\end{equation}
so that $q_c(t)=r_c(t)/z(t)$.

We next examine one small update interval. Let
$\Delta g_c(t)=g_c(t+\Delta t)-g_c(t)$. Since
\begin{equation}
\tilde{w}_c(t+\Delta t)
=
\tilde{w}_c(t)\exp(-\Delta g_c(t)),
\end{equation}
the first-order expansion $\exp(-\Delta g_c(t))=1-\Delta g_c(t)+o(\Delta t)$ gives
\begin{equation}
\Delta r_c(t)
=
r_c(t+\Delta t)-r_c(t)
=
-r_c(t)\Delta g_c(t)+o(\Delta t).
\end{equation}
Similarly,
\begin{equation}
\Delta z(t)
=
\sum_{j=1}^C \Delta r_j(t)
=
-\sum_{j=1}^C r_j(t)\Delta g_j(t)+o(\Delta t).
\end{equation}
Using $q_c(t)=r_c(t)/z(t)$, the change of the normalized class probability is
\begin{equation}
\begin{aligned}
\Delta q_c(t)
&=
\frac{r_c(t)+\Delta r_c(t)}
{z(t)+\Delta z(t)}
-
\frac{r_c(t)}{z(t)}\\
&=
\frac{z(t)\Delta r_c(t)-r_c(t)\Delta z(t)}
{z(t)\left(z(t)+\Delta z(t)\right)}
+o(\Delta t).
\end{aligned}
\end{equation}
Since $\Delta z(t)$ is small, $z(t)(z(t)+\Delta z(t))=z^2(t)+o(\Delta t)$. Substituting the expressions of $\Delta r_c(t)$ and $\Delta z(t)$ yields
\begin{equation}
\begin{aligned}
\Delta q_c(t)
&=
-q_c(t)\Delta g_c(t)\\
&\quad
+q_c(t)\sum_{j=1}^C q_j(t)\Delta g_j(t)+o(\Delta t)\\
&=
-q_c(t)\left[
\Delta g_c(t)-\sum_{j=1}^C q_j(t)\Delta g_j(t)
\right]+o(\Delta t).
\end{aligned}
\end{equation}
Taking the continuous-time limit gives the class-level differential form
\begin{equation}
dq_c(t)
=
-q_c(t)\left[
dg_c(t)-\sum_{j=1}^C q_j(t)dg_j(t)
\right].
\end{equation}

It remains to model the stochastic evolution of $g_c(t)$. In SGS, $g_c(t)$ is updated from mini-batch frequency and sharpness statistics. These estimates contain sampling noise, especially for sparse tail classes. We therefore use the continuous-time stochastic approximation
\begin{equation}
dg_c(t)=\mu_c(t)dt+\sigma_c(t)\circ dW_c(t),
\end{equation}
where $\mu_c(t)$ is the deterministic drift of the frequency-curvature drive, $\sigma_c(t)$ is the volatility induced by mini-batch estimation, and $W_c(t)$ is a standard Wiener process. The Stratonovich interpretation is adopted because the SDE is obtained as the limit of smooth discrete updates, which preserves the ordinary chain rule used above. Substitution gives
\begin{equation}
\begin{aligned}
dq_c(t)
&=
-q_c(t)\left[
\mu_c(t)-\sum_{j=1}^C q_j(t)\mu_j(t)
\right]dt\\
&\quad
-q_c(t)\sigma_c(t)\circ dW_c(t)\\
&\quad
+q_c(t)\sum_{j=1}^C q_j(t)\sigma_j(t)\circ dW_j(t).
\end{aligned}
\end{equation}

Finally, consider an instance $z_i=(x_i,y_i)$. Under intra-class uniformity, its instance-level probability is $q_i(t)=q_{y_i}(t)/N_{y_i}$. Since $N_{y_i}$ is constant, $dq_i(t)=dq_{y_i}(t)/N_{y_i}$. Replacing $c$ with $y_i$ in the class-level SDE and defining
\begin{equation}
\bar{\mu}(t)=\sum_{c=1}^C q_c(t)\mu_c(t)
\end{equation}
gives
\begin{equation}
\begin{aligned}
dq_i(t)
&=
-q_i(t)\left[
\mu_{y_i}(t)-\bar{\mu}(t)
\right]dt\\
&\quad
-q_i(t)\sigma_{y_i}(t)\circ dW_{y_i}(t)\\
&\quad
+q_i(t)\sum_{c=1}^C q_c(t)\sigma_c(t)\circ dW_c(t),
\end{aligned}
\end{equation}
which is the statement of Theorem~1.
\end{proof}

\subsection{Proof of Theorem 2}
\begin{proof}
We focus on the class-level sampling probability of a fixed class $c$ and write
$q=q_c(t)$. From Theorem~1, the class-level dynamics are
\begin{equation}
\begin{aligned}
dq_c(t)
&=-q_c(t)\left[\mu_c(t)-\bar{\mu}(t)\right]dt\\
&\quad -q_c(t)\sigma_c(t)\circ dW_c(t)\\
&\quad +q_c(t)\sum_{j=1}^C q_j(t)\sigma_j(t)\circ dW_j(t),
\end{aligned}
\end{equation}
where $\bar{\mu}(t)=\sum_{j=1}^C q_j(t)\mu_j(t)$. In the local time window used for stationary analysis, we freeze the slowly varying coefficients and write
\begin{equation}
a_c=\mu_c-\bar{\mu}.
\end{equation}
The diffusion part can be written as
\begin{equation}
\begin{aligned}
&-q\sigma_c\circ dW_c(t)
+q\sum_{j=1}^C q_j\sigma_j\circ dW_j(t)\\
&\quad =
q\left[
(q-1)\sigma_c\circ dW_c(t)
+\sum_{j\ne c}q_j\sigma_j\circ dW_j(t)
\right].
\end{aligned}
\end{equation}
Assuming independent Brownian increments, this linear combination has local variance
\begin{equation}
\tilde{\sigma}_c^2
=(1-q)^2\sigma_c^2+\sum_{j\ne c}q_j^2\sigma_j^2 .
\end{equation}
Thus it can be represented by an effective Brownian motion $W_t$, with the sign absorbed into $W_t$, and the reduced Stratonovich SDE becomes
\begin{equation}
dq=-a_cq\,dt+q\tilde{\sigma}_c\circ dW_t .
\end{equation}
The finite sampler assigns positive probability to every class. We therefore analyze the process on a compact interval $q\in[q_{\min},q_{\max}]\subset(0,1)$ and impose zero probability flux at the boundaries.

Let
\begin{equation}
A(q)=-a_cq,\qquad B(q)=q\tilde{\sigma}_c .
\end{equation}
To use the standard Fokker--Planck equation, we convert the Stratonovich SDE to its It\^o form. The It\^o drift is
\begin{equation}
A_{\mathrm{It\hat{o}}}(q)
=A(q)+\frac{1}{2}B(q)B'(q)
=-a_cq+\frac{1}{2}q\tilde{\sigma}_c^2,
\end{equation}
and the diffusion coefficient is
\begin{equation}
D(q)=B^2(q)=q^2\tilde{\sigma}_c^2.
\end{equation}
The density $\varphi(q,t)$ of the It\^o process satisfies the Fokker--Planck equation
\begin{equation}
\frac{\partial \varphi(q,t)}{\partial t}
=
-\frac{\partial}{\partial q}
\left[A_{\mathrm{It\hat{o}}}(q)\varphi(q,t)\right]
+
\frac{1}{2}\frac{\partial^2}{\partial q^2}
\left[D(q)\varphi(q,t)\right].
\end{equation}
At stationarity, $\partial\varphi(q,t)/\partial t=0$. Integrating once with respect to $q$ gives a constant probability flux
\begin{equation}
J(q)
=
A_{\mathrm{It\hat{o}}}(q)\varphi(q)
-
\frac{1}{2}\frac{d}{dq}\left[D(q)\varphi(q)\right].
\end{equation}
The zero-flux boundary condition sets $J(q)=0$, hence
\begin{equation}
A_{\mathrm{It\hat{o}}}(q)\varphi(q)
=
\frac{1}{2}\frac{d}{dq}\left[D(q)\varphi(q)\right].
\end{equation}
Substituting
$A_{\mathrm{It\hat{o}}}(q)=-a_cq+\frac{1}{2}q\tilde{\sigma}_c^2$
and $D(q)=q^2\tilde{\sigma}_c^2$ gives
\begin{equation}
\left(-a_cq+\frac{1}{2}q\tilde{\sigma}_c^2\right)\varphi(q)
=
\frac{1}{2}\frac{d}{dq}
\left[q^2\tilde{\sigma}_c^2\varphi(q)\right].
\end{equation}
Expanding the derivative on the right-hand side,
\begin{equation}
\frac{1}{2}\frac{d}{dq}
\left[q^2\tilde{\sigma}_c^2\varphi(q)\right]
=
q\tilde{\sigma}_c^2\varphi(q)
+
\frac{1}{2}q^2\tilde{\sigma}_c^2\varphi'(q).
\end{equation}
Therefore,
\begin{equation}
-a_cq\varphi(q)+\frac{1}{2}q\tilde{\sigma}_c^2\varphi(q)
=
q\tilde{\sigma}_c^2\varphi(q)
+
\frac{1}{2}q^2\tilde{\sigma}_c^2\varphi'(q).
\end{equation}
Rearranging terms,
\begin{equation}
\frac{1}{2}q^2\tilde{\sigma}_c^2\varphi'(q)
=
-a_cq\varphi(q)
-
\frac{1}{2}q\tilde{\sigma}_c^2\varphi(q).
\end{equation}
Dividing both sides by
$\frac{1}{2}q^2\tilde{\sigma}_c^2\varphi(q)$ yields the logarithmic derivative
\begin{equation}
\frac{\varphi'(q)}{\varphi(q)}
=
-\frac{1}{q}
\left(
1+\frac{2a_c}{\tilde{\sigma}_c^2}
\right).
\end{equation}
Integrating with respect to $q$ gives
\begin{equation}
\log \varphi(q)
=
-\left(
1+\frac{2a_c}{\tilde{\sigma}_c^2}
\right)\log q
+
C_c .
\end{equation}
Here $C_c$ is the integration constant for class $c$. Taking the exponential and absorbing $\exp(C_c)$ into the normalizer, we obtain
\begin{equation}
\varphi_{\infty}^{(c)}(q)
=
\frac{1}{z_c}q^{-\gamma_c},
\qquad
\gamma_c
=
1+\frac{2a_c}{\tilde{\sigma}_c^2}
=
1+\frac{2(\mu_c-\bar{\mu})}{\tilde{\sigma}_c^2},
\end{equation}
where
\begin{equation}
z_c=\int_{q_{\min}}^{q_{\max}}q^{-\gamma_c}dq
\end{equation}
normalizes the density on the compact positive-probability support.

The exponent $\gamma_c$ determines the direction of the stationary shift. If $\mu_c<\bar{\mu}$, then $\gamma_c$ decreases and the density is less concentrated near the lower boundary, so the class retains larger exposure. If $\mu_c>\bar{\mu}$, then $\gamma_c$ increases and the density shifts toward smaller probabilities, suppressing over-exposed or sharper classes. Thus the stationary distribution balances class exposure and geometry rather than performing naive tail over-sampling. This proves the theorem.
\end{proof}

\subsection{Proof of Theorem 3}
\begin{proof}
Let $\delta\in(0,e^{-1}]$, and assume the sharp loss is bounded in $[0,1]$. Let $\mathbf{q}(t)=(q_1(t),\ldots,q_C(t))$ be the class-level sampling distribution at time $t$, and let $\mathbf{u}=(1/C,\ldots,1/C)$. For class $c$, let $S_c(t)$ denote the possibly reweighted effective class-$c$ sample induced by the sampler, and write $n_c(t)=nq_c(t)>1$ for its effective sample size. Define
\begin{equation}\label{equ:appendix_class_sharp_risk}
\begin{aligned}
\mathcal{L}_c(w)
&=\mathbb{E}_{z\sim\mathcal{D}_c}[\ell(w,z)],\\
\widehat{\mathcal{L}}_{c,S}^{\rho}(w)
&=\frac{1}{n_c(t)}\sum_{z_i\in S_c(t)}
\max_{\|\epsilon\|_2\leq\rho}\ell(w+\epsilon,z_i),
\end{aligned}
\end{equation}
where $\mathcal{D}_c$ is the class-conditional data distribution. The sampling-weighted sharp empirical risk is
\begin{equation}\label{equ:appendix_q_sharp_risk}
\widehat{\mathcal{L}}_{\mathbf{q}(t),S}^{\rho}(w)
=\sum_{c=1}^C q_c(t)\widehat{\mathcal{L}}_{c,S}^{\rho}(w).
\end{equation}
Under intra-class uniform sampling, the sampler first chooses a class according to $\mathbf{q}(t)$ and then samples uniformly inside that class; hence the instance-level sharp empirical risk is represented by $\widehat{\mathcal{L}}_{\mathbf{q}(t),S}^{\rho}(w)$.

Apply the SAM PAC-Bayes bound~\cite{foretsharpness} separately to each effective class sample $S_c(t)$. The only change from the original SAM bound is that the sample size becomes $n_c(t)$; assigning confidence $\delta_c=\delta/C$ to each class and using a union bound makes the result hold for all classes at once. Thus, with probability at least $1-\delta$, for all $c$,
\begin{equation}\label{equ:appendix_class_pac}
\mathcal{L}_c(w)
\leq
\widehat{\mathcal{L}}_{c,S}^{\rho}(w)
+
\sqrt{
\frac{
\mathrm{KL}(\mathcal{P}_w\|\mathcal{P}_0)+\log(C n_c(t)/\delta)}
{2(n_c(t)-1)}
},
\end{equation}
where $\mathcal{P}_w$ is the posterior and $\mathcal{P}_0$ is the prior.

Since $\mathcal{L}_{bal}(w)=\frac{1}{C}\sum_{c=1}^C\mathcal{L}_c(w)$, averaging \eqref{equ:appendix_class_pac} gives
\begin{equation}\label{equ:appendix_balanced_average}
\begin{aligned}
\mathcal{L}_{bal}(w)
&\leq
\frac{1}{C}\sum_{c=1}^C\widehat{\mathcal{L}}_{c,S}^{\rho}(w)\\
&\quad+
\frac{1}{C}\sum_{c=1}^C
\sqrt{
\frac{
\mathrm{KL}(\mathcal{P}_w\|\mathcal{P}_0)+\log(C n_c(t)/\delta)}
{2(n_c(t)-1)}
}.
\end{aligned}
\end{equation}

The first term in \eqref{equ:appendix_balanced_average} is a uniform class average, while training under $\mathbf{q}(t)$ observes a $\mathbf{q}(t)$-weighted average. Let $r_c=\widehat{\mathcal{L}}_{c,S}^{\rho}(w)$. By the bounded sharp-loss assumption, $r_c\in[0,1]$,
\begin{equation}\label{equ:appendix_mismatch_bound}
\begin{aligned}
\frac{1}{C}\sum_{c=1}^C r_c
&=
\sum_{c=1}^C q_c(t)r_c
+
\sum_{c=1}^C\left(\frac{1}{C}-q_c(t)\right)r_c\\
&\leq
\widehat{\mathcal{L}}_{\mathbf{q}(t),S}^{\rho}(w)
+
\sum_{c=1}^C\left|\frac{1}{C}-q_c(t)\right|\\
&=
\widehat{\mathcal{L}}_{\mathbf{q}(t),S}^{\rho}(w)
+
\|\mathbf{q}(t)-\mathbf{u}\|_1 .
\end{aligned}
\end{equation}
Combining \eqref{equ:appendix_balanced_average}--\eqref{equ:appendix_mismatch_bound} with $n_c(t)=nq_c(t)$ gives
\begin{equation}\label{equ:appendix_final_pac}
\begin{aligned}
\mathcal{L}_{bal}(w)
&\leq
\widehat{\mathcal{L}}_{\mathbf{q}(t),S}^{\rho}(w)
+
\|\mathbf{q}(t)-\mathbf{u}\|_1\\
&\quad+
\frac{1}{C}\sum_{c=1}^{C}
\sqrt{
\frac{
\mathrm{KL}(\mathcal{P}_w\|\mathcal{P}_0)+\log(Cnq_c(t)/\delta)}
{2(nq_c(t)-1)}
} .
\end{aligned}
\end{equation}

For vanilla long-tailed SAM, the sampler follows the empirical prior $q_c(t)\equiv\pi_c=N_c/N$, so
\begin{equation}\label{equ:appendix_vanilla_lt_bound}
\begin{aligned}
\mathcal{L}_{bal}(w)
&\leq
\widehat{\mathcal{L}}_{\boldsymbol{\pi},S}^{\rho}(w)
+
\|\boldsymbol{\pi}-\mathbf{u}\|_1\\
&\quad+
\frac{1}{C}\sum_{c=1}^{C}
\sqrt{
\frac{
\mathrm{KL}(\mathcal{P}_w\|\mathcal{P}_0)+\log(Cn\pi_c/\delta)}
{2(n\pi_c-1)}
}.
\end{aligned}
\end{equation}
This bound is loose in long-tailed data for two reasons: $\|\boldsymbol{\pi}-\mathbf{u}\|_1$ measures the mismatch between empirical sampling and balanced evaluation, and small tail $\pi_c$ makes the complexity denominator $2(n\pi_c-1)$ small.

For any class-level sampling distribution $\mathbf{q}$ with $nq_c>1$ for all classes, collect the sampling-dependent part of the bound as
\begin{equation}\label{equ:appendix_sampling_penalty}
\begin{aligned}
\mathcal{B}(\mathbf{q})
&=
\|\mathbf{q}-\mathbf{u}\|_1\\
&\quad+
\frac{1}{C}\sum_{c=1}^{C}
\sqrt{
\frac{
\mathrm{KL}(\mathcal{P}_w\|\mathcal{P}_0)+\log(Cnq_c/\delta)}
{2(nq_c-1)}
}.
\end{aligned}
\end{equation}
We first consider the second term of $\mathcal{B}(\mathbf{q})$, which depends on the effective class sample size. Let
\begin{equation}
\Phi(q)=
\sqrt{
\frac{
\mathrm{KL}(\mathcal{P}_w\|\mathcal{P}_0)+\log(Cnq/\delta)}
{2(nq-1)}
}.
\end{equation}
Write $A(q)=\mathrm{KL}(\mathcal{P}_w\|\mathcal{P}_0)+\log(Cnq/\delta)$. For $q>1/n$,
\begin{equation}
\Phi'(q)
=
\frac{\Phi(q)}{2}
\left(
\frac{1}{qA(q)}
-
\frac{n}{nq-1}
\right).
\end{equation}
Since $\mathrm{KL}(\mathcal{P}_w\|\mathcal{P}_0)\geq0$, $C\geq1$, and $nq>1$, we have
\begin{equation}
A(q)\geq \log(Cnq/\delta)>\log(1/\delta).
\end{equation}
For the high-probability PAC setting $\delta\leq e^{-1}$, this implies $A(q)>1>1-\frac{1}{nq}$, and hence $\Phi'(q)<0$. Thus increasing the effective exposure of a class reduces its class-wise complexity contribution.

For the full class average, we assume that the movement from the empirical prior $\boldsymbol{\pi}$ to the stationary class-level distribution $\mathbf{q}^{\star}=(q_1^{\star},\ldots,q_C^{\star})$ admits an equalizing-transfer decomposition. Each transfer moves probability from a class $j$ with larger probability to a class $i$ with smaller probability, so $q_i<q_j$, and transfers an amount $\varepsilon$ without crossing. We further assume that the transfer occurs in the effective-sample region where $\Phi$ is decreasing and locally strictly convex. Then $\Phi'$ is strictly increasing on the transfer interval, and
\begin{equation}
\begin{aligned}
&\Phi(q_i+\varepsilon)+\Phi(q_j-\varepsilon)
-\Phi(q_i)-\Phi(q_j)\\
&=
\int_{0}^{\varepsilon}
\left[\Phi'(q_i+s)-\Phi'(q_j-s)\right]ds
<0 .
\end{aligned}
\end{equation}
This shows that each assumed equalizing transfer decreases the pairwise complexity sum: the reduction on the lower-probability class dominates the increase on the higher-probability class. Applying this argument to all transfers in the decomposition gives the following complexity-side comparison:
\begin{equation}\label{equ:appendix_theorem2_complexity}
\frac{1}{C}\sum_{c=1}^{C}\Phi(q_c^{\star})
<
\frac{1}{C}\sum_{c=1}^{C}\Phi(\pi_c).
\end{equation}

We next consider the first term of $\mathcal{B}(\mathbf{q})$. Theorem~2 motivates a shift from classes with larger frequency-curvature drive to classes with smaller drive. To compare the mismatch term rigorously, we assume that the resulting stationary sampler moves closer to the class-balanced prior:
\begin{equation}\label{equ:appendix_stationary_balance_assumption}
\|\mathbf{q}^{\star}-\mathbf{u}\|_1
<
\|\boldsymbol{\pi}-\mathbf{u}\|_1 .
\end{equation}
This assumption does not require exact class-uniform sampling; it only states that SGS reduces the mismatch between the training sampler and the balanced evaluation prior. Combining \eqref{equ:appendix_theorem2_complexity} with \eqref{equ:appendix_stationary_balance_assumption} gives
\begin{equation}\label{equ:appendix_sampling_penalty_reduced}
\mathcal{B}(\mathbf{q}^{\star})
<
\mathcal{B}(\boldsymbol{\pi}) .
\end{equation}
It remains to compare the full PAC-Bayes right-hand side, which also contains the sharp empirical risk. The sampling-dependent penalty has been reduced by \eqref{equ:appendix_sampling_penalty_reduced}, but the sharp empirical term under $\mathbf{q}^{\star}$ may differ from that under $\boldsymbol{\pi}$. We therefore use the net-improvement condition
\begin{equation}\label{equ:appendix_net_improvement}
\widehat{\mathcal{L}}_{\mathbf{q}^{\star},S}^{\rho}(w)
-
\widehat{\mathcal{L}}_{\boldsymbol{\pi},S}^{\rho}(w)
<
\mathcal{B}(\boldsymbol{\pi})
-
\mathcal{B}(\mathbf{q}^{\star}) .
\end{equation}
This condition allows the sharp empirical term to increase, as long as that increase is smaller than the reduction in the sampling-dependent penalty. Under \eqref{equ:appendix_net_improvement}, the SGS right-hand side satisfies
\begin{equation}\label{equ:appendix_sges_tighter}
\begin{aligned}
&\widehat{\mathcal{L}}_{\mathbf{q}^{\star},S}^{\rho}(w)
+
\mathcal{B}(\mathbf{q}^{\star})\\
&<
\widehat{\mathcal{L}}_{\boldsymbol{\pi},S}^{\rho}(w)
+
\left[\mathcal{B}(\boldsymbol{\pi})
-
\mathcal{B}(\mathbf{q}^{\star})\right]
+
\mathcal{B}(\mathbf{q}^{\star})\\
&=
\widehat{\mathcal{L}}_{\boldsymbol{\pi},S}^{\rho}(w)
+
\mathcal{B}(\boldsymbol{\pi}) .
\end{aligned}
\end{equation}
Thus, under the bounded sharp-loss, effective-sample, equalizing-transfer, stationary-balance, and net-improvement assumptions above, the SGS stationary sampler has a strictly smaller PAC-Bayes upper bound than direct long-tailed SAM without requiring exact class-uniform sampling. This proves Theorem~3.
\end{proof}

\section{More Experiment Protocols}
\label{app:more_experiment_protocols}

\subsection{Datasets}
\begin{itemize}
    \item \textbf{CIFAR-10 LT and CIFAR-100 LT}~\cite{cao2019learning}. These datasets are constructed by exponentially reducing the number of training samples per class from the original balanced CIFAR datasets. We report results under imbalance ratios $\mathrm{IR}\in\{10,50,100,200\}$, where IR denotes the ratio between the largest and smallest class sizes.
    \item \textbf{ImageNet-LT}~\cite{liu2019large}. ImageNet-LT contains 1,000 categories, 115,846 training images, and 50,000 validation images. The dataset is constructed from ImageNet with an imbalance ratio of 256.
    \item \textbf{iNaturalist}~\cite{van2018inaturalist}. iNaturalist is a naturally long-tailed fine-grained recognition dataset with 8,142 categories and a substantially larger imbalance ratio.
\end{itemize}

\subsection{Evaluation Protocol}
\begin{itemize}
    \item \textbf{Classification metric.} We report top-1 accuracy on balanced test sets. When subset results are available, classes are grouped into Head, Medium, and Tail splits following common long-tailed recognition practice.
    \item \textbf{CIFAR-10 LT split.} For CIFAR-10 LT with $\mathrm{IR}=100$, Head classes contain more than 1,000 samples, Medium classes contain 200--1,000 samples, and Tail classes contain fewer than 200 samples.
    \item \textbf{CIFAR-100 LT, ImageNet-LT, and iNaturalist splits.} For these datasets, Head classes contain more than 100 samples, Medium classes contain 20--100 samples, and Tail classes contain fewer than 20 samples.
    \item \textbf{Other imbalance ratios.} For CIFAR-LT under $\mathrm{IR}\in\{10,50,200\}$, we report the overall top-1 accuracy.
\end{itemize}

\subsection{Implementation Details}
\begin{itemize}
    \item \textbf{CIFAR-LT training.} We use ResNet-32~\cite{he2016deep} and train for 200 epochs with SGD, momentum 0.9, batch size 64, and a cosine learning-rate schedule.
    \item \textbf{ImageNet-LT and iNaturalist training.} For ResNet models, we use ResNet-50 and train for 200 epochs with SGD and momentum 0.9. For ImageNet-LT, the batch size is 256.
    \item \textbf{Deferred LA fine-tuning.} We first train SGS-SAM for 180 epochs, then freeze the feature extractor and fine-tune only the classifier for 20 epochs with LA and SAM.
    \item \textbf{Foundation-model fine-tuning.} We follow the CLIP/ViT protocol of Focal-SAM and LIFT. Specifically, we use CLIP~\cite{radford2021learning} with a ViT-B/16 image encoder~\cite{dosovitskiyimage}. A cosine classifier is attached after the image encoder and initialized using CLIP text features; the text encoder is discarded during fine-tuning. The CLIP image encoder is fine-tuned for 20 epochs with SGD, batch size 128, and momentum 0.9.
    \item \textbf{SGS hyperparameters.} Unless otherwise specified, we set the sharpness exponent to $\kappa=0.6$ according to the hyperparameter analysis in the main text. The warm-up ratio $r_w$ is selected according to the imbalance ratio: more severe imbalance uses earlier SGS activation, while milder imbalance uses a longer warm-up stage.
    \item \textbf{Reported statistics and speed.} When reporting standard deviations, we average over multiple runs with different random seeds. To reduce the influence of hardware and implementation differences, the training-speed comparison reports relative ratios with respect to vanilla SAM rather than absolute running time.
\end{itemize}

\section{More Experiment Results}
\label{app:More_Experiment_Results}

\subsection{Sampling Distribution Analysis}
\label{app:sampling_distribution}

We directly examine whether SGS changes the realized training exposure rather than only the nominal sampling weights. The comparison is conducted on CIFAR-100 LT with $\mathrm{IR}=100$ for 200 epochs and a batch size of 64. For SAM, we count the class labels of the indices returned by the original shuffled long-tailed batch sampler. For SGS-SAM, we count the indices actually returned by the dynamic frequency-sharpness sampler during training. Both methods use the same dataset split. 

\begin{center}
  \centering
  \includegraphics[width=\textwidth]{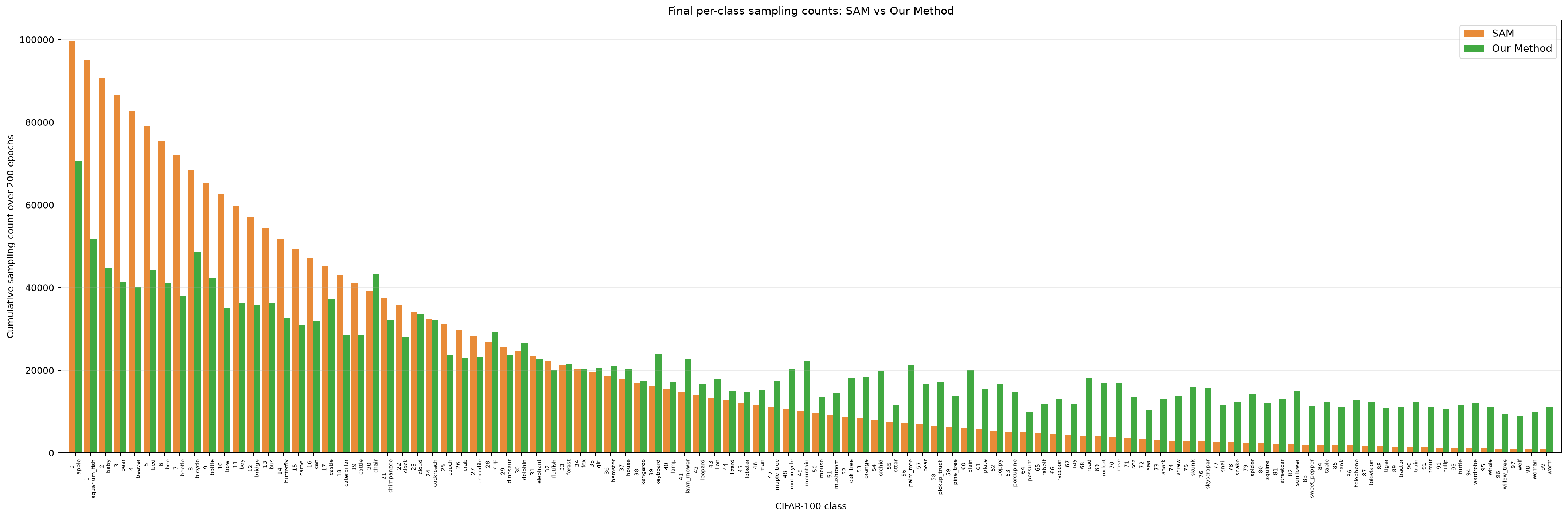}
  \captionof{figure}{Cumulative per-class sampling counts on CIFAR-100 LT with $\mathrm{IR}=100$. Classes are ordered by the original long-tailed class frequency. ``Our Method'' in the figure denotes SGS-SAM. Counts are obtained from the indices actually returned by each batch sampler over 200 epochs.}
  \label{fig:appendix_sampling_distribution}
\end{center}

Figure~\ref{fig:appendix_sampling_distribution} shows that SAM largely preserves the original long-tailed exposure profile: the most frequent head classes are sampled tens of thousands of times more often than the rarest tail classes. SGS-SAM substantially compresses this range by reducing repeated head-class draws and increasing medium- and tail-class exposure. The resulting profile is more balanced, but not exactly uniform, because SGS retains part of the original sampler and further modulates each class by its online sharpness estimate. Consequently, classes with similar frequencies may receive different counts when their perturbation responses differ.

This observation is consistent with the stochastic dynamics developed in the main paper. In Theorem~1, the relative drift $-q_c(\mu_c-\bar{\mu})$ transfers sampling mass toward classes with lower frequency-sharpness drive and suppresses classes whose drive is above the population average. The head-to-tail compression in Fig.~\ref{fig:appendix_sampling_distribution} is the finite-horizon empirical counterpart of this feedback. It also supports the stationary interpretation in Theorem~2: SGS approaches a frequency-curvature equilibrium rather than a purely class-uniform distribution. Finally, the more balanced exposure makes $\widehat{\mathbf{q}}^{\mathrm{SGS}}$ closer to the balanced prior than the empirical SAM exposure, increasing the effective sample sizes of tail classes.

\subsection{Hessian Spectrum Analysis}
\label{app:hessian_spectrum}

The maximum Hessian eigenvalue is commonly used as a sharpness and flatness metric~\cite{keskar-2016-large_batch-ICLR,jastrzkebski2017three,chaudhari-2019-Entropy_sgd-IOP}.
We next examine how the loss geometry relates to the generalization bound in Theorem~3. For class $c$, let $H_c=\nabla^2\mathcal{L}_c(\theta)$ denote the class-wise Hessian at a solution $\theta$. Its largest eigenvalue $\lambda_{\max}(H_c)$ measures the sharpest local curvature direction. A smaller $\lambda_{\max}$ therefore indicates a flatter minimum, provides geometric evidence for a smaller sharp empirical-risk term in Theorem~3, and is commonly associated with stronger generalization~\cite{foretsharpness}. We estimate the Hessian eigenvalue spectral density using ResNet-32 on CIFAR-10 LT with $\mathrm{IR}=100$ under CE training. Following Focal-SAM~\cite{li2025focal}, we report the spectra for Head, Medium, Tail, and All classes.
We employ Hutchinson's method~\cite{yao2020pyhessian} to estimate the spectrum.

\begin{center}
  \centering
  \begin{minipage}{0.49\textwidth}
    \centering
    \includegraphics[width=\linewidth]{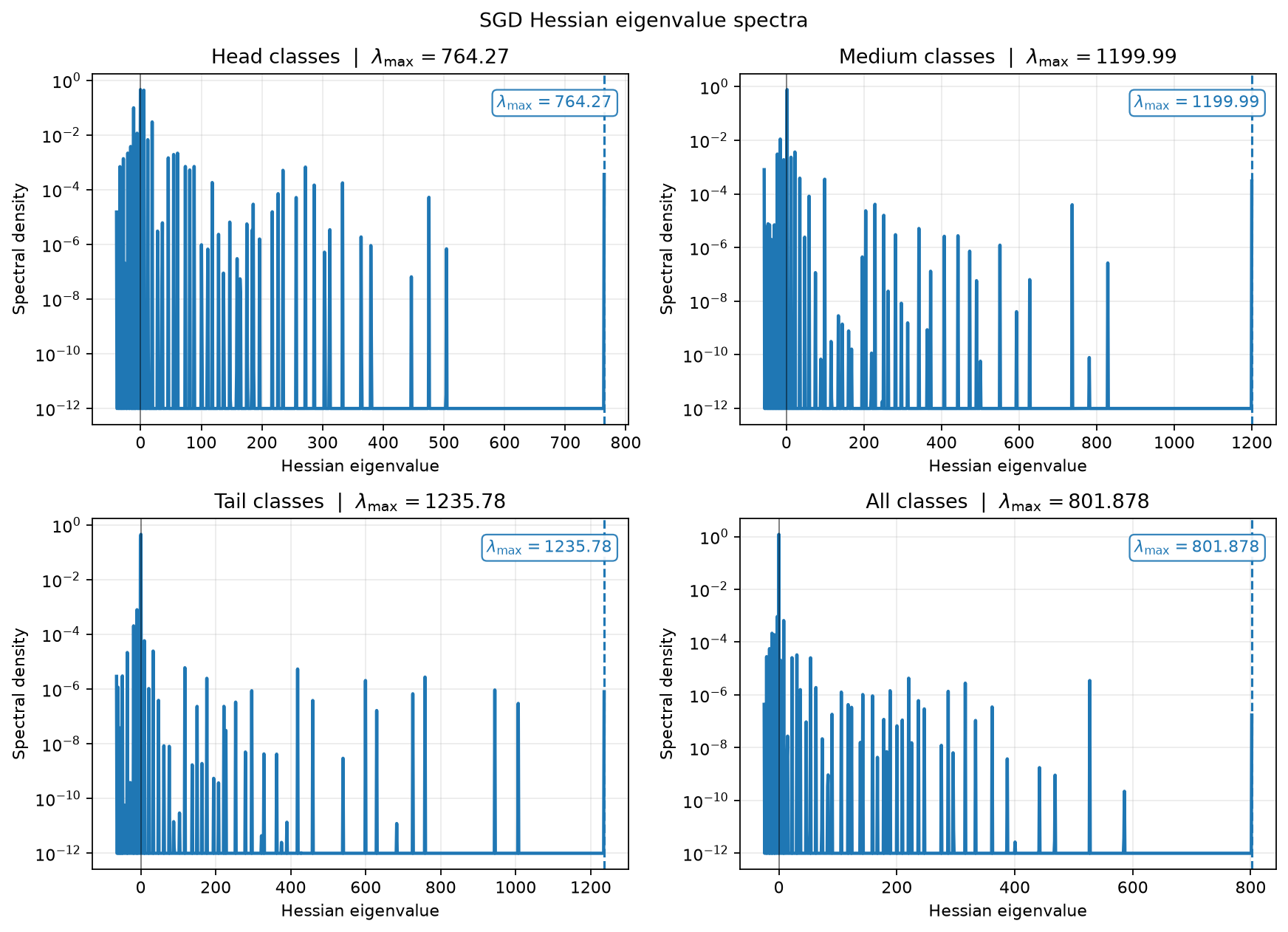}\\
    \small (a) SGD
  \end{minipage}
  \hfill
  \begin{minipage}{0.49\textwidth}
    \centering
    \includegraphics[width=\linewidth]{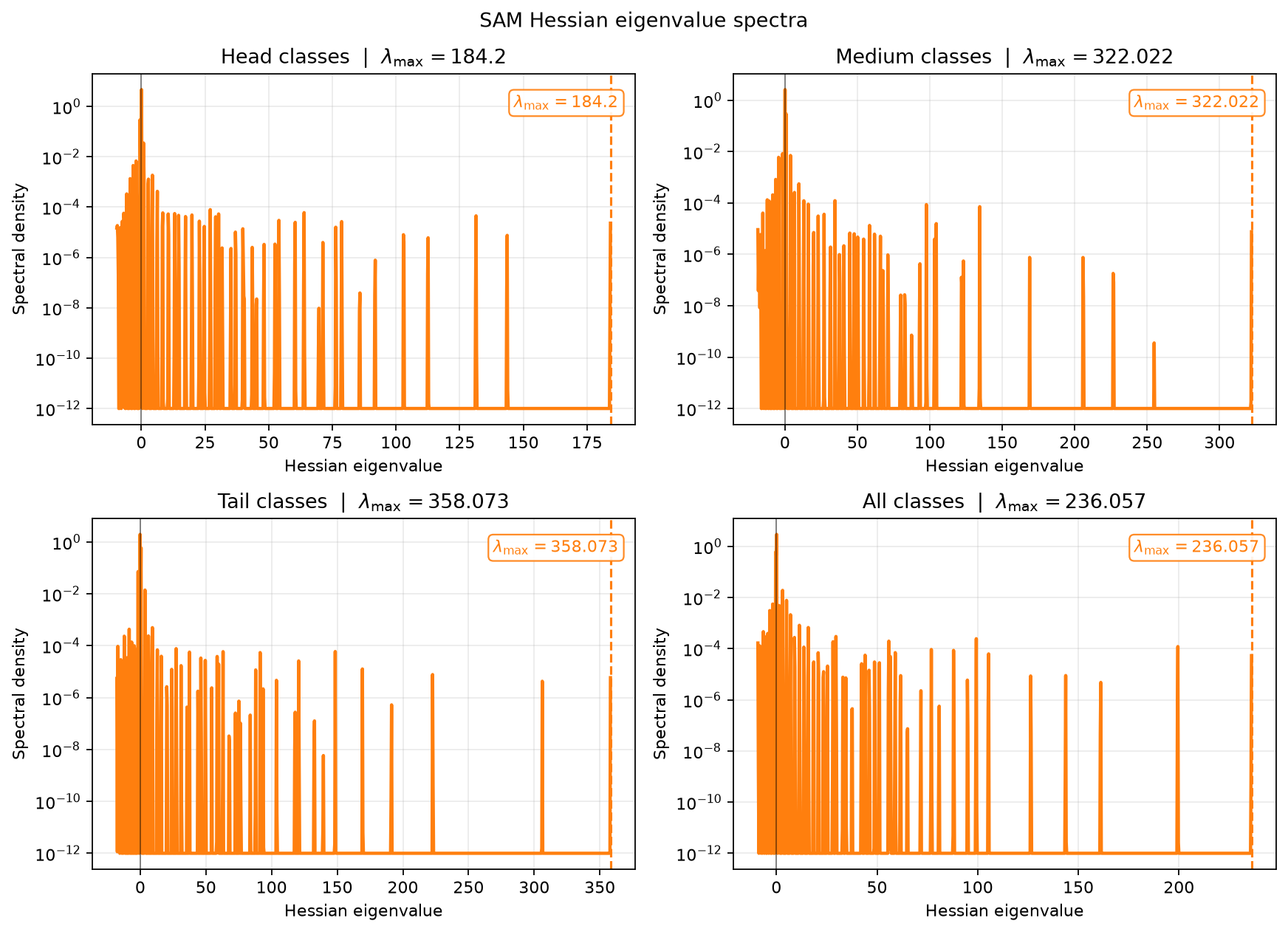}\\
    \small (b) SAM
  \end{minipage}

  \begin{minipage}{0.49\textwidth}
    \centering
    \includegraphics[width=\linewidth]{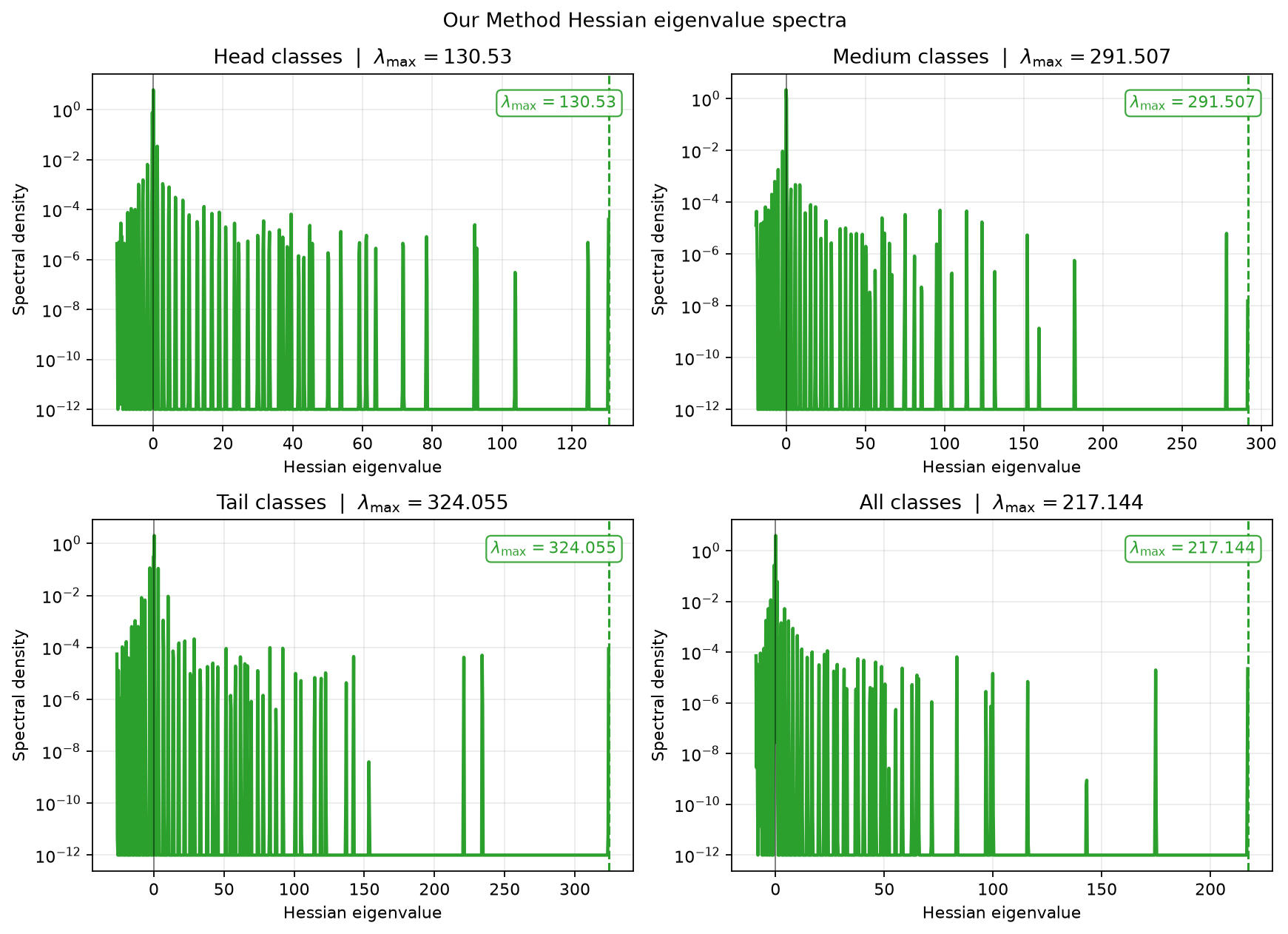}\\
    \small (c) SGS-SAM
  \end{minipage}
  \captionof{figure}{Hessian eigenvalue spectral density for Head, Medium, Tail, and All classes on CIFAR-10 LT with $\mathrm{IR}=100$. The dashed line marks $\lambda_{\max}$. A smaller $\lambda_{\max}$ indicates a flatter loss landscape.}
  \label{fig:appendix_hessian_spectrum}
\end{center}

As shown in Fig.~\ref{fig:appendix_hessian_spectrum}, SGD exhibits a long positive spectral tail, with $\lambda_{\max}$ values of 764.27, 1199.99, 1235.78, and 801.878 for Head, Medium, Tail, and All classes, respectively. SAM substantially contracts the spectral edge to 184.2, 322.022, 358.073, and 236.057. SGS-SAM further reduces these values to 130.53, 291.507, 324.055, and 217.144, corresponding to reductions of 29.14\%, 9.48\%, 9.50\%, and 8.01\% over SAM. The improvement is observed across all class groups and is largest for Head classes, while the Medium and Tail spectra are also consistently compressed. These results show that frequency-sharpness-guided sampling complements the SAM perturbation: by adjusting training exposure before each update, SGS-SAM reduces sharp directions without sacrificing the flatness of another class group, yielding a more uniformly flat loss geometry across the long-tailed distribution.

\subsection{Component Ablation}
\label{app:component_ablation}

We isolate the contributions of the two sampling indicators on CIFAR-100 LT with $\mathrm{IR}=100$ using ResNet-32 and CE training. We denote the variant using only $w_c^{\mathrm{freq}}$ as \emph{SGS-F}, and the variant using only $w_c^{\mathrm{sharp}}$ as \emph{SGS-S}. For SGS-S, we set $\kappa=1$ so that the inverse relative sharpness is used without exponent smoothing. All other training settings follow the main experiment.

\begin{center}
  \begin{minipage}{\textwidth}
  \centering
  \captionof{table}{Component ablation on CIFAR-100 LT with $\mathrm{IR}=100$ using ResNet-32. SGS-F and SGS-S denote frequency-only and sharpness-only sampling, respectively. The SAM and full SGS-SAM results are those reported in the main paper.}
  \setlength{\tabcolsep}{4.5pt}
  \begin{tabular}{@{}lccccccc@{}}
  \toprule
  Method & $w^{\mathrm{freq}}$ & $w^{\mathrm{sharp}}$ & $\kappa$ & Head & Med & Tail & All \\
  \midrule
  SAM & -- & -- & -- & 72.70 & 41.80 & 7.00 & 42.20 \\
  SGS-F & \checkmark & -- & -- & 72.24 & 46.03 & 17.79 & 46.53 \\
  SGS-S & -- & \checkmark & 1 & 72.62 & 38.53 & 5.17 & 40.19 \\
  \textbf{SGS-SAM} & \checkmark & \checkmark & 0.6 & \textbf{72.97} & \textbf{46.58} & \textbf{19.75} & \textbf{47.56} \\
  \bottomrule
  \end{tabular}
  \label{tab:appendix_component_ablation}
  \end{minipage}
\end{center}

Table~\ref{tab:appendix_component_ablation} shows that the frequency indicator provides the main rebalancing effect. Compared with SAM, SGS-F improves medium, tail, and overall accuracy by 4.23, 10.79, and 4.33 points, respectively, while preserving comparable head accuracy. SGS-S alone instead obtains 40.19 overall accuracy. Tail classes are often both under-exposed and locally unstable because their sharpness estimates are formed from fewer and noisier samples. Without the frequency indicator, inverse-sharpness sampling may suppress these high-response tail classes rather than compensating for their insufficient exposure. It can therefore favor classes that are already locally stable while leaving the long-tailed sampling mismatch unresolved, which explains the lower medium and tail accuracies of SGS-S.

Combining both indicators gives the best result in every column. Relative to SGS-F, full SGS-SAM further improves head, medium, tail, and overall accuracy by 0.73, 0.55, 1.96, and 1.03 points, respectively. Thus, $w_c^{\mathrm{freq}}$ restores effective exposure for under-represented classes, while $w_c^{\mathrm{sharp}}$ regulates this redistribution using local loss geometry.

\subsection{Hessian Analysis of the Ablated Samplers}
We further study the generalization-relevant geometry produced by SGS-F and SGS-S. Following the Hessian protocol above, we train ResNet-32 on CIFAR-10 LT with $\mathrm{IR}=100$ under CE and plot the class-wise spectra of both variants.

\begin{center}
  \centering
  \begin{minipage}{0.49\textwidth}
    \centering
    \includegraphics[width=\linewidth]{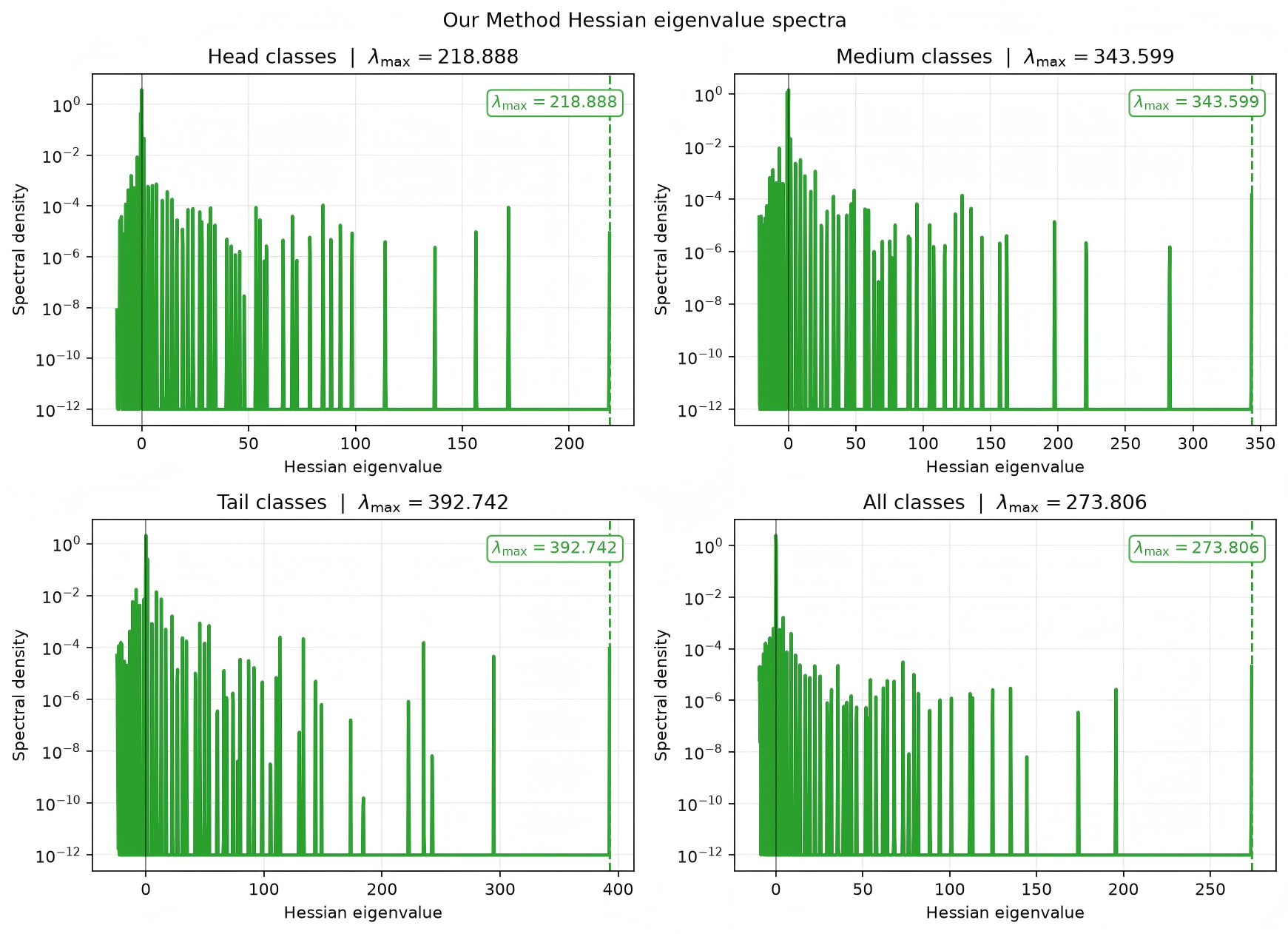}\\
    \small (a) SGS-F
  \end{minipage}
  \hfill
  \begin{minipage}{0.49\textwidth}
    \centering
    \includegraphics[width=\linewidth]{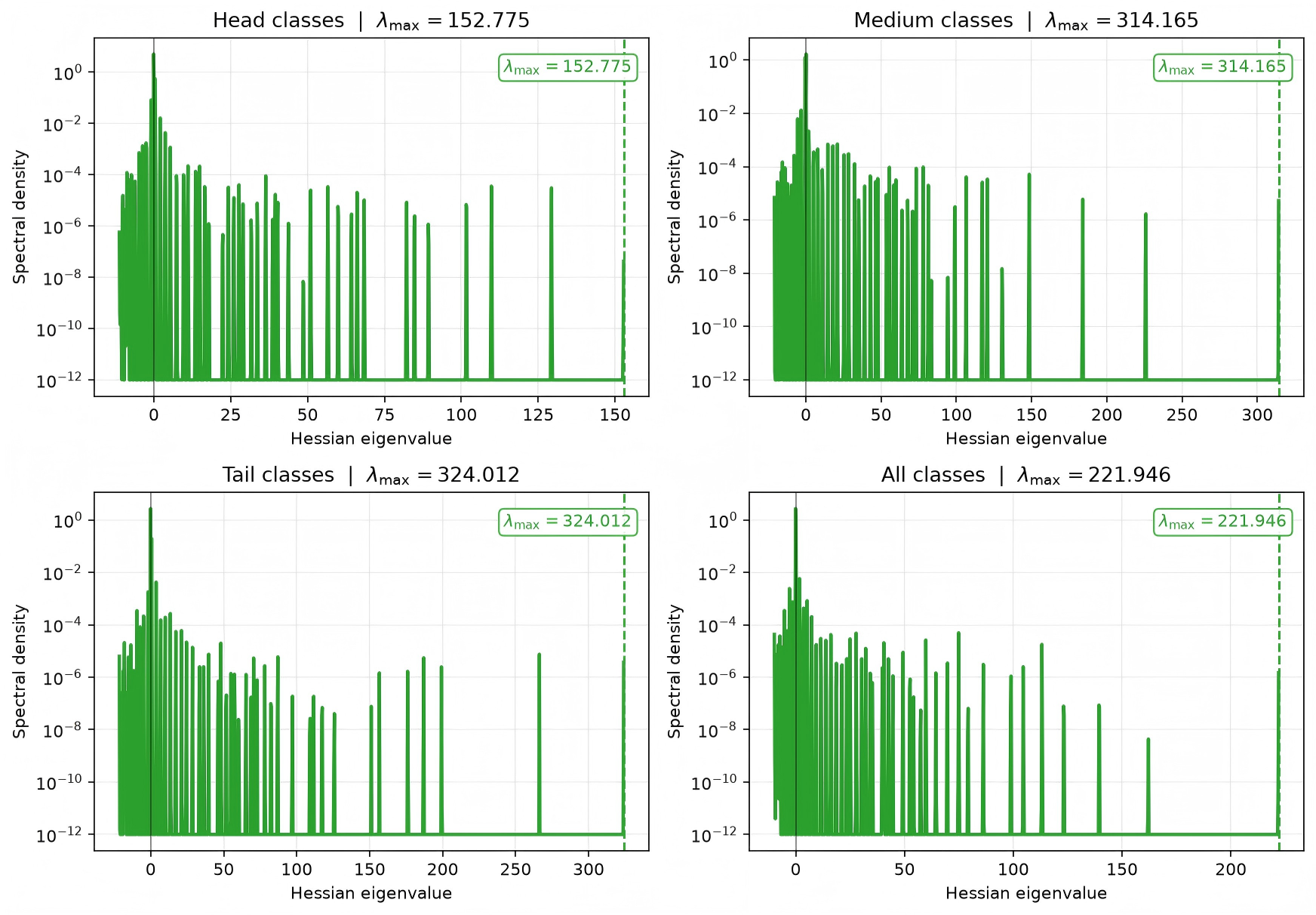}\\
    \small (b) SGS-S
  \end{minipage}
  \captionof{figure}{Hessian eigenvalue spectral density of the ablated samplers using ResNet-32 on CIFAR-10 LT with $\mathrm{IR}=100$. SGS-F uses only $w_c^{\mathrm{freq}}$, whereas SGS-S uses only $w_c^{\mathrm{sharp}}$ with $\kappa=1$.}
  \label{fig:appendix_component_hessian}
\end{center}

Figure~\ref{fig:appendix_component_hessian} reveals different geometric effects. SGS-F obtains $\lambda_{\max}$ values of 218.888, 343.599, 392.742, and 273.806 for Head, Medium, Tail, and All classes, respectively. These values are 18.83\%, 6.70\%, 9.68\%, and 15.99\% higher than those of SAM. Frequency-only rebalancing repeatedly increases the exposure of scarce classes but cannot reduce their sampling rate when the perturbation response becomes excessive. Consequently, noisy or high-curvature updates remain unregulated, so SGS-F improves balanced accuracy through greater tail exposure while converging to a sharper solution.

In contrast, SGS-S obtains $\lambda_{\max}$ values of 152.775, 314.165, 324.012, and 221.946, reducing the corresponding SGS-F values by 30.20\%, 8.57\%, 17.50\%, and 18.94\%. The sharpness feedback directly suppresses repeated sampling from locally unstable regions and steers optimization toward flatter neighborhoods. However, the accuracy results show that flatness alone is insufficient under class imbalance: without frequency correction, SGS-S does not guarantee adequate tail exposure. This explains why SGS-S has the flatter spectrum but the lower classification accuracy, whereas SGS-F has better tail accuracy but a sharper spectrum.

The complete SGS-SAM combines these effects. Relative to SGS-F, it reduces the Head, Medium, Tail, and All spectral edges by 40.37\%, 15.16\%, 17.49\%, and 20.69\%, while also achieving the best classification accuracy in Table~\ref{tab:appendix_component_ablation}. The frequency indicator addresses sampling mismatch and effective sample size, whereas the sharpness indicator prevents this redistribution from being dominated by unstable high-curvature updates. This result establishes the necessity of incorporating sharpness into the sampling process: sharpness serves not only as a diagnostic of the final solution, but also as actionable feedback that shapes future data exposure. More broadly, SGS shows that a sampler can jointly control statistical balance and optimization geometry, extending re-sampling from static class correction to geometry-aware training control.

\subsection{Training-Loss Dynamics}
\label{app:training_loss_dynamics}

We further compare the training-loss trajectories of SGD, SAM, SGS-F, SGS-S, and the complete SGS-SAM under the CIFAR-100 LT ablation setting above. These curves complement the final accuracy and Hessian results by showing when each sampling signal begins to affect optimization.

\begin{center}
  \centering
  \includegraphics[width=0.95\textwidth]{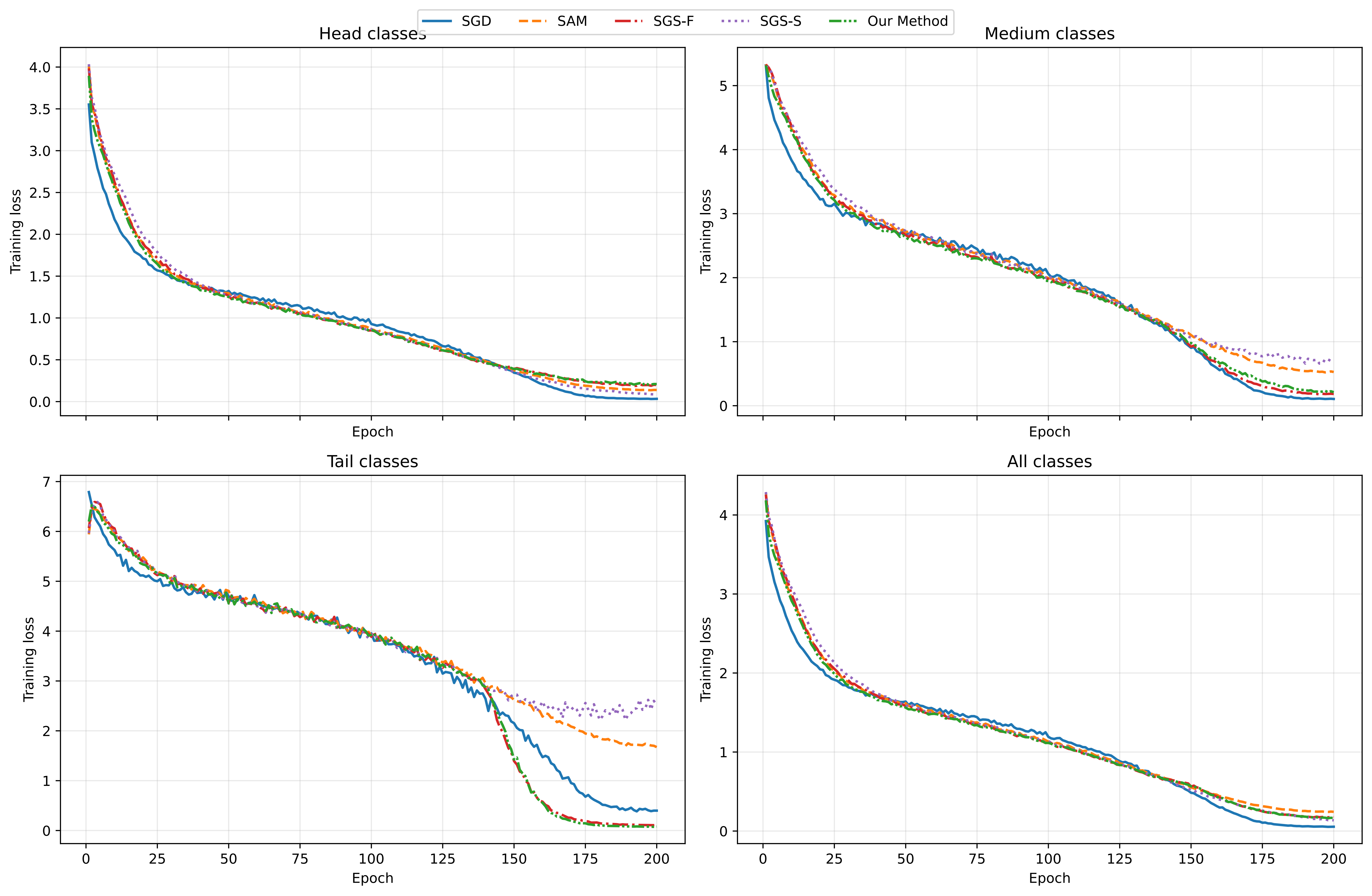}
  \captionof{figure}{Training loss on Head, Medium, Tail, and All classes for SGD, SAM, SGS-F, SGS-S, and SGS-SAM on CIFAR-100 LT with $\mathrm{IR}=100$ using ResNet-32. ``Our Method'' in the legend denotes the complete SGS-SAM.}
  \label{fig:appendix_training_loss}
\end{center}

Figure~\ref{fig:appendix_training_loss} shows that the methods have similar early and middle-stage trends, while their tail-class behavior separates late in training. The tail loss of SGS-F and SGS-SAM drops rapidly after approximately 145 epochs and approaches zero by the end of training, whereas SAM and SGS-S retain substantially larger tail losses. This confirms that the frequency indicator needs sufficient cumulative updates before its additional tail exposure is translated into optimization progress. SGS-S remains flatter in the Hessian analysis but has the highest final tail loss because sharpness-only sampling can suppress unstable tail classes without compensating for their low frequency. SGS-F fits the tail aggressively, but its larger Hessian eigenvalues indicate that this reduction is obtained in a sharper region. SGS-SAM retains the late tail-loss reduction of SGS-F while using sharpness feedback to obtain the flatter geometry and better balanced accuracy reported above. The slightly lower final loss of SGD or SGS-F on some groups does not contradict this result, because training loss alone does not measure class-balanced generalization or neighborhood flatness.

This delayed tail improvement also explains why SGS-SAM requires a longer adaptation schedule on iNaturalist. The dataset contains 8,142 classes, so many tail classes occur only sporadically in a mini-batch; consequently, the class-wise frequency counts and EMA sharpness estimates require more iterations to become reliable. Moreover, the foundation-model protocol in the main paper sets $r_w=0$ but limits $\alpha_{\max}$ to $0.1$ to preserve the long-tailed prior used by LA. Thus, only a small fraction of each mini-batch is drawn from the SGS distribution, and the corrective tail exposure accumulates slowly even without a formal warm-up stage. A short 20-epoch schedule may terminate before many rare classes receive enough SGS-controlled updates. Longer training allows the online statistics to stabilize and gives the redistributed samples sufficient time to reduce tail loss, making the benefit of geometry-aware sampling more fully observable on iNaturalist.

\subsection{Instance Segmentation Results}
\textbf{Dataset and Evaluation.} We additionally evaluate SGS on long-tailed instance segmentation using the LVIS protocol of Balanced Meta-SoftMax~\cite{ren2020balanced}. LVIS~\cite{gupta2019lvis} contains 1,230 categories with severe class imbalance. We report COCO-style AP~\cite{lin2014microsoft}: mask AP over all/frequent/common/rare classes ($AP_m$, $AP_f$, $AP_c$, $AP_r$) and bounding-box AP ($AP_b$).

\textbf{Architectures and Training.} We conduct the experiment on top of the Balanced Meta-SoftMax protocol~\cite{ren2020balanced}. Specifically, we use the released ImageNet-pretrained Mask R-CNN baseline with a ResNet-50 backbone and apply SGS in the decoupled training stage. Following BALMS, decoupled training refers to training the last linear classifier on a fixed feature extractor obtained from instance-balanced training. The detection-head classifier is fine-tuned for 22k iterations under the same SGD settings, augmentation, and single-scale testing protocol as BALMS. During this stage, SGS accumulates foreground RoI frequency and sharpness statistics by category, while background RoIs retain the standard detector loss.

\begin{table}[h]
  \centering
  \caption{Additional results on LVIS instance segmentation. All compared results are from Balanced Meta-SoftMax~\cite{ren2020balanced}.}
  \setlength{\tabcolsep}{2pt}
  \begin{tabular}{@{}lccccc@{}}
  \toprule
  Method & $AP_m$ & $AP_f$ & $AP_c$ & $AP_r$ & $AP_b$ \\
  \midrule
  Softmax & 23.7 & 27.3 & 24.0 & 13.6 & 24.0 \\
  Sigmoid & 23.6 & 27.3 & 24.0 & 12.7 & 24.0 \\
  Focal Loss & 23.4 & 27.5 & 23.5 & 12.8 & 23.8 \\
  Class-Balanced Loss & 23.3 & 27.3 & 23.8 & 11.4 & 23.9 \\
  LDAM & 24.1 & 26.3 & 25.3 & 14.6 & 24.5 \\
  LWS & 23.8 & 26.8 & 24.4 & 14.4 & 24.1 \\
  Equalization Loss & 25.2 & 26.6 & 27.3 & 14.6 & 25.7 \\
  Balanced Softmax$^\dagger$ & 26.3 & \textbf{28.8} & 27.3 & 16.2 & 27.0 \\
  BALMS & 27.0 & 27.5 & 28.9 & 19.6 & 27.6 \\
  \textbf{SGS-SAM} & 25.12 & 28.07 & 26.15 & 14.52 & 24.39 \\
  \bottomrule
  \end{tabular}
  \label{tab:appendix_lvis_instance_segmentation}
\end{table}

SGS-SAM improves over several baselines, including Sigmoid, Focal Loss, Class-Balanced Loss, LDAM, and LWS on overall mask AP, and obtains the best frequent-class AP among the listed methods. However, it remains below BALMS on overall mask AP and rare-category AP. This limitation is expected because long-tailed instance segmentation contains imbalance not only across images or categories, but also within each image: different foreground RoIs, object sizes, and background regions form an internal long-tailed structure. The current SGS sampler mainly controls the external training exposure through image/category-level statistics, so it cannot fully correct the fine-grained RoI-level imbalance inside a detector. These results suggest that SGS transfers reasonably to decoupled detector-head training, while segmentation may require an RoI-aware or detector-specific extension.

\bibliography{aaai2026}
